\documentclass[twoside,11pt]{article}
\usepackage{jair, theapa, rawfonts}

\usepackage[ruled,vlined]{algorithm2e}
\usepackage{amsmath, multirow, amssymb, mathrsfs, dsfont}
\usepackage{booktabs}
\usepackage{graphicx}
\usepackage{longtable}
\usepackage{afterpage}
\usepackage{subfigure}
\usepackage{color}
\usepackage[normalem]{ulem}

\allowdisplaybreaks

\DeclareMathOperator*{\argmin}{\arg\,min}
\DeclareMathOperator*{\argmax}{\arg\,max}

\jairheading{78}{2023}{167-215}{09/2022}{10/2023}
\ShortHeadings{Amortized Variational Inference: A Systematic Review}
{Ganguly, Jain, \& Watchareeruetai}
\firstpageno{167}

\begin{document}

\title{Amortized Variational Inference: A Systematic Review}

\author{\name Ankush Ganguly  \email agang@sertiscorp.com \\
       \name Sanjana Jain \email sjain@sertiscorp.com \\
       \name Ukrit Watchareeruetai \email uwatc@sertiscorp.com \\
       \addr Sertis Vision Lab\\
      Sukhumvit Road, Watthana, Bangkok 10110, Thailand
}


\maketitle

\begin{abstract}
The core principle of Variational Inference (VI) is to convert the statistical inference problem of computing complex posterior probability densities into a tractable optimization problem. 
This property enables VI to be faster than several sampling-based techniques.
However, the traditional VI algorithm is not scalable to large data sets and is unable to readily infer out-of-bounds data points without re-running the optimization process.
Recent developments in the field, like stochastic-, black box-, and amortized-VI, have helped address these issues.
Generative modeling tasks nowadays widely make use of amortized VI for its efficiency and scalability, as it utilizes a parameterized function to learn the approximate posterior density parameters.
In this paper, we review the mathematical foundations of various VI techniques to form the basis for understanding amortized VI.
Additionally, we provide an overview of the recent trends that address several issues of amortized VI, such as the amortization gap, generalization issues, inconsistent representation learning, and posterior collapse. 
Finally, we analyze alternate divergence measures that improve VI optimization.
\end{abstract}

\section{Introduction}
\label{Introduction}

Bayesian inference is an indispensable part of machine learning as it allows for systematic reasoning about parameter uncertainty \shortcite{zhang_vi_2019}.
Bayesian statistics' core principle is to frame all inference about unknown variables as a calculation involving a posterior probability density \shortcite{blei2017variational}.
Exact inference, which typically involves analytically computing the posterior probability distribution over the variables of interest, offers a solution to this inference problem.
Algorithms in this category include the elimination algorithm \shortcite{cozman2000generalizing}, the sum-product algorithm \shortcite{Kschischang2001Factor}, and the junction tree algorithm \shortcite{madsen1999lazy}.
For highly complex probability densities, however, exact inference does not guarantee a closed-form solution. 
In fact, the exact computation of conditional probabilities in belief networks is NP-hard \shortcite{dagum_luby_1993}.

As an alternative to exact inference, approximate inference techniques, which have been in development since the early 1950s, offer an efficient solution to Bayesian inference by providing simpler estimates of complex probability densities.
Approximate inference provides solutions to even non-conjugate\footnote{Conjugacy occurs when the posterior density is in the same family of probability density functions as the prior, but with new parameter values which have been updated to reflect the learning from the data.} models for which analytic posteriors are unavailable \shortcite{knollmuller2019metric}.
Markov Chain Monte Carlo (MCMC) methods such as the Metropolis-Hastings algorithm \shortcite{metropolis1953} and Gibbs sampling \shortcite{geman_1987} fall under this category.
However, MCMC methods that rely on sampling \shortcite{brooks2011handbook,gershman2012nonparametric,Robbins_1951} are slow to converge and do not scale efficiently.

Variational Inference (VI), a method in machine learning, tackles the problem of inefficient approximate inference by the use of a suitable metric to select a tractable approximation to the posterior probability density.
The methodology of VI is, thus, to re-frame the statistical inference problem into an optimization problem giving us the speed benefits of maximum a posteriori (MAP) estimation \shortcite{Murphy1991} and the ability to scale to large data sets \shortcite{blei2017variational}.
This makes VI an ideal choice for application in areas like statistical physics \shortcite{regier2015celeste,smith_2021,marino_manic_2021}, diagnostic inference in Quick Medical Reference (QMR) networks \shortcite{jakkola_qmr}, generative modeling \shortcite<e.g.,>{kingma2013auto,larsen2016autoencoding,zhao_song_ermon_2019,higgins2016beta,burgess2018understanding}, and neural networks \shortcite<e.g.,>{sun2019functional,shen_chen_deng_2020,haussmann2020sampling,eikema_aziz_2019}.
Other than VI, loopy-belief propagation \shortcite{murphy2013loopy} and expectation maximization \shortcite{minka2013expectation} also fall within the class of optimization-based inference techniques.
We re-iterate that both MCMC methods and VI solve the problem of inference, but their respective approaches are different. While MCMC algorithms rely on sampling to approximate the posterior, VI uses optimization for the approximation.

Since its inception, researchers have developed the traditional VI algorithm \shortcite<introduced by>{jordan1999introduction} to make it more accurate, efficient, and scalable.
The traditional VI algorithm operates by introducing a new set of parameters, characterizing the approximated density, for every observation with the aim to find unbiased estimates for the parameters of the true posterior probability density.
This leads to inefficient scalability as these optimizable parameters grow linearly with the observations.
On the other hand, \textit{amortized inference}, an improvement over the traditional VI algorithm, uses a stochastic function to estimate the posterior probability density \shortcite{zhang_vi_2019}.
Unlike traditional VI, the parameters of this stochastic function are fixed and shared across all data points, thereby amortizing the inference\footnote{Section \ref{section:avi} explains in detail the intent of the use of this phrase in the context of probabilistic modeling.}.
Deep neural networks are a popular choice for this stochastic function as they combine probabilistic modeling with the representational power of deep learning \shortcite{zhang_vi_2019}.
Thus, amortized inference combined with deep neural networks has been shown to efficiently scale to large data sets.
The variational auto-encoder (VAE) \shortcite{kingma2013auto,RezendeMW14} and its variants are primary examples in this case.
Not only does amortizing the inference aid in scalability, but the memoized re-use \shortcite{gershman2014amortized} of the learned parameters of the stochastic function helps test inference on new observations without having to re-run the optimization process, like in the case of traditional VI.

In this paper, we study and provide an intuitive explanation of the different VI techniques and their applications to researchers new to the field, and elucidate the strengths and weaknesses of these methods. In addition, this paper builds off of the mathematical foundations of traditional-, stochastic-, and black box-VI to form the basis of explanation for amortized VI, its properties, and caveats.
To the best of our knowledge, while several excellent reviews of VI exist \shortcite<e.g.,>{blei2017variational,zhang_vi_2019}, this is the first review paper dedicated to gaining a deeper understanding of the concept of amortized VI while distinguishing it from several other forms of VI.
In addition, we unify the mathematical notations from many research papers to ease the readers in understanding the concepts, features, and differences of each VI methodology. 
Furthermore, we study how the recent developments in the field of amortized VI have addressed its weaknesses, discuss alternate divergence measures, and analyze their effect on VI optimization.

We organize the paper as follows:
Section \ref{section:vi} discusses the relevance of statistical inference in real-world problems as well as revisits the core concepts of VI such as the VI optimization problem, the Evidence Lower Bound (ELBO), mean field VI, and the coordinate ascent VI (CAVI) optimization algorithm.
Section \ref{section:svi} explains how stochastic VI uses stochastic optimization and natural gradients to make VI a scalable method.
Section \ref{section:bbvi} elucidates the concept of black box VI and the reparameterization trick.
Section \ref{section:avi} dives into a deeper understanding of amortized inference, addresses the issues associated with it, and provides an overview of the various advancements.
Section \ref{section:beyond} analyzes the use of different divergence measures that improve optimization in VI.
Finally, we conclude with an overview of active research areas and open problems in Section \ref{section:open}.

\begin{table}[t!]
\centering
\begin{tabular}{@{} l l}
\toprule
Notation        & Description \\
\midrule
$x = [x_{1},..., x_{N}]$       & Observed variable \\
$z = [z_{1},..., z_{N}]$       & Latent variable \\
$N$    & Total number of data points  \\
$\theta$    & Global generative model parameters\\
$\phi$    & Global variational (or recognition model) parameters\\
$\xi_{i}$    & Local variational parameter for the $i$-th data point\\
$p(z|x_{i}; \theta)$    & True posterior probability density conditioned on $\theta$ \\
$q(z|x_{i}; \xi_{i})$    & Variational approximation parameterized by $\xi_{i}$ \\
$\mathcal{Q}$    & Family of tractable probability densities\\
$D_{\text{KL}}(q(z|x; \xi)\;\|\;p(z|x; \theta))$    & Kullback-Leibler divergence \\
$\mathcal{D}$ & The total KL-divergence objective function \\
$M$    & Mini-batch size for stochastic optimization \\
$x^{M}$ & Mini-batch containing $M$ data points \\
$\xi^{M}$ & Set of the local variational parameters for $M$ samples\\
$\mathcal{M}$ & The set of all the subsets of the data set split into $\frac{N}{M}$ subsets \\
$\mathcal{L}$    & Evidence Lower Bound (ELBO)\\
$\mathcal{\hat{L}}$    & Stochastic estimate of the ELBO \\
$\mathcal{\Tilde{L}}$    & Stochastic estimate of the ELBO using Monte Carlo samples \\
$\mathcal{\Tilde{L}}^{B}$ & SGVB estimator for VAE \shortcite{kingma2013auto} \\
$K$ & Monte Carlo samples used in ELBO approximation \\
$\nabla$ & Stochastic gradients of a function \\
$\hat{\nabla}$ & Mean stochastic gradients of a function based on $M$ samples\\
$\Bar{\nabla}$ & The natural gradients of a function \\
$I$ & The Fisher Information Matrix \\
\bottomrule
\end{tabular}
\caption{Notations used throughout this paper.}
\par
\label{tab:notations}
\end{table}

\section{Variational Inference}
\label{section:vi}

This section explores the importance of statistical inference in practical problems and reviews core VI concepts like the VI optimization problem, ELBO, mean field VI, and the CAVI optimization algorithm.
\subsection{Statistical Inference in Real-world Problems}
\label{susbsection:pgm}

Statistical inference has been applied to solve many real-world problems. 
As an example, let's consider the problem of topic modeling \shortcite{lda_2003,online_lda_2010,hoffmanstochastic,topic_model_2012} that aims to uncover hidden thematic structures, i.e., topics, within each document in a corpus. 
A topic in a document refers to a subject that represents a meaningful pattern of words found in the document. 
It represents a set of words that are frequently associated with each other within a specific context. 
A document can be considered as a mixture of topics, where a distribution over words characterizes each topic. 
In topic modeling, the words in the documents are called observable variables since they can be directly observed from the documents, while topics, the underlying variables influencing the distributions of words in documents, are called latent variables.
These latent variables can be inferred using statistical inference.

Another example of a real-world problem is a recommender system, which aims to predict the preferences of a user on a particular product. 
Collaborative filtering is one of the major approaches for recommender systems. 
In collaborative filtering, a user-item interaction matrix, in which rows and columns represent users and items, respectively, and each element in the matrix denotes the preference, e.g., rating score, of a particular user on a particular item. 
This matrix is, however, sparse and incomplete; the values of most elements are unknown. 
Collaborative filtering aims to fill in this incomplete matrix by assuming that there is a group of latent variables, called a latent vector, affecting the observed values in each row and each column. 
Once latent vectors of all users and items are obtained, the user-item interaction matrix can be simply computed using matrix multiplication.  Similarly, statistical inference is used to obtain the parameters of a generative process to produce these latent variables \cite{collab_vae_2017,vae_collab_filtering_2018}.

\begin{figure}[htbp]
  \centering
  \subfigure[Without plate notation]{
    \includegraphics[width=0.45\textwidth]{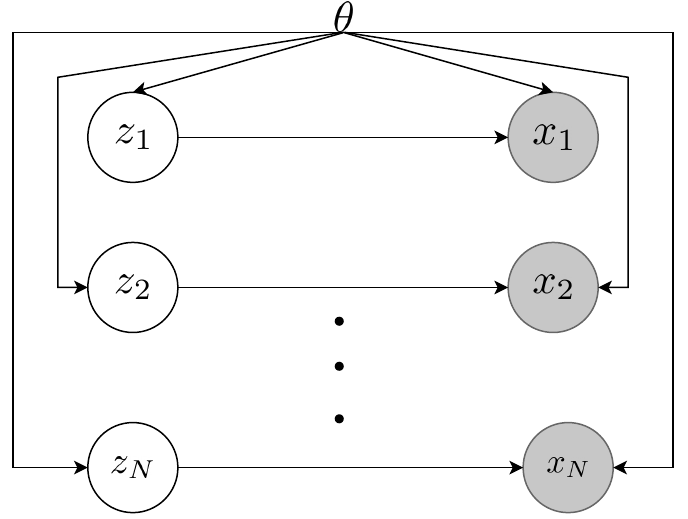}
    \label{fig:wo_plate_notation}
  }
  \hfill
  \subfigure[With plate notation]{
    \includegraphics[width=0.45\textwidth]{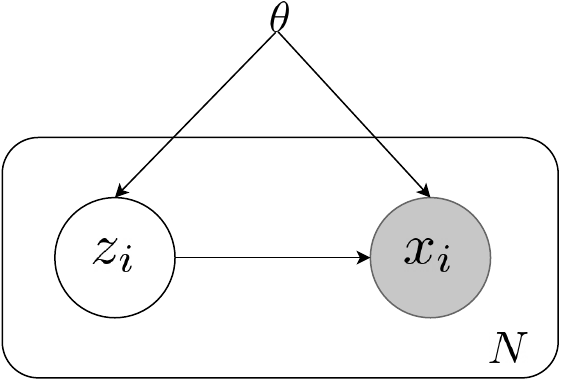}
    \label{fig:with_plate_notation}
  }
  
  \caption{A directed graphical model with $N$ data points describing joint distributions over $x_i$ and $z_i$ as follows: $p(x_i, z_i; \theta) = p(x_i | z_i; \theta) p(z_i ; \theta)$.}
  \label{fig:side_by_side}
\end{figure}

Graphical model representation \shortcite{getting_started_pgm,graphical_model,bishop_pattern_recognit_pgm} is a tool allowing us to formulate a real-world problem as a graph that can be solved by statistical inference.
A graphical model is a probabilistic model that expresses conditional dependence structure between random variables using a graph.
Figure \ref{fig:wo_plate_notation} is an example of a graphical model.
In a graphical model, a node can represent either a random variable or a deterministic parameter.
A random variable is represented by a circle while a deterministic parameter is denoted by just a symbol.
For example, as shown in the figure, there are two groups of random variables, i.e., $x_i$ and $z_i$ where $i=1,\dots,N$, and a deterministic parameter $\theta$.
An edge, i.e., a link between two nodes, denotes the conditional dependence between them; for example, the edge from $\theta$ to $z_i$ indicates that each random variable $z_i$ is conditioned on $\theta$.
Moreover, a graphical model allows one to differentiate observed variables from latent variables by shading the corresponding nodes.
From the figure, $x_i$ are observed variables while $z_i$ are latent variables.
When a node is enclosed by a plate (denoted by the rounded-corner rectangle in Figure \ref{fig:with_plate_notation}), it indicates that there is a group of nodes of the same kind.

Our assumption is that the observed data points are independent and identically distributed (i.i.d.) and are generated by some random process, involving the unobserved random variable $z$.
For each data point $x_{i}$, there exists a latent vector $z_{i}$, which is assumed to have some prior probability density $p(z; \theta)$. 
Additionally, the data points are sampled from the conditional probability density $p(x|z; \theta)$.
Computing the posterior probability density, $p(z|x; \theta)$, is useful for a variety of tasks such as coding or data representation, denoising, recognition, and visualization \shortcite{kingma2013auto}.

Although a variety of generative processes with more complex directed graphical models are possible, we restrict ourselves to the common case where a latent variable is associated with each i.i.d. observed data point.
For example, in our case, the observed data points can be pictured as images while the latent variables as lower dimensional representations of those images.
From a coding theory perspective, these latent variables are interpreted as \textit{code} \shortcite{kingma2013auto} and thus form the basis of representation learning.

We use Bayes' theorem to compute the posterior probability density as:
\begin{align}
    \label{equation:bayes}
     p(z|x; \theta) &= \frac{p(x|z; \theta)p(z; \theta)}{p(x; \theta)},\nonumber\\
     p(x;\theta) &=\int p(x|z; \theta)p(z; \theta) \mathrm{d}z.
\end{align}
The marginal probability density, $p(x; \theta)$, in Equation \ref{equation:bayes} is called the \textit{evidence}, which is high dimensional for most statistical models, and its computation is thus, at times, intractable or of exponential complexity. 
This computation is significant as a higher marginal likelihood indicates the chosen model's ability to fit the observed data better.

The purpose of VI is, therefore, two-fold:
\begin{enumerate}
    \item Analytical approximation of the posterior probability density for statistical inference over the latent variables.
    \item Provide an alternative to tractably compute the evidence to encourage a better fit to the data by the chosen statistical model.
\end{enumerate}

\subsection{Statistical Inference as Optimization}
\label{susbsection:kld}

\begin{figure}[th]
    \centering
    \includegraphics[clip, width=0.4\columnwidth]{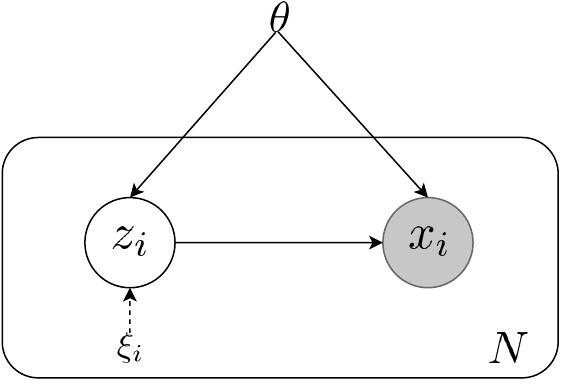}
    \caption{A directed graphical model with $N$ data points. Solid lines denote the generative model, while dashed lines denote the variational approximation to the intractable posterior density \shortcite{kingma2013auto}.
    The local variational parameters and the global generative model parameters are represented by $\xi_{i}$ and $\theta$, respectively.
    }
    \label{fig:pgm}
\end{figure}

The central idea of VI is to provide simpler approximations to complex posterior probability densities.
Traditionally, for each data point $x_{i}$, VI aims to select the approximate density, $q(z|x_{i}; \xi_{i})$, from a family of tractable densities $\mathcal{Q}$ (Figure \ref{fig:pgm}).
Each $q(z|x_{i}; \xi_{i}) \in \mathcal{Q}$ is designated by a set of their own variational parameters, $\xi_{i}$,  and is a candidate approximation of the actual posterior evaluated at data point $x_{i}$.
The goal is to tune these parameters to get an optimal approximation of the actual posterior density.
Thus, VI converts this Bayesian inference problem into an optimization problem.
The complexity and the accuracy of this optimization are controlled by the choice of the variational family \shortcite{plummer2020dynamics}.
This choice, further, depends on a measure that captures the difference between the approximated posterior and the true posterior density \shortcite{ranganath2014black}.
Usually, this measure is chosen to be the non-negative Kullback–Leibler (KL) divergence which estimates the relative entropy between two densities \shortcite{10.5555/1162264,kullback_leibler_1951,jordan1999introduction}.
In the case of VI, it quantifies the relative entropy between the true posterior probability density, $p(z|x_{i}; \theta)$, and the candidate density, $q(z|x_{i}; \xi_{i})$.
The optimization problem for traditional VI entails reducing the relative entropy by choosing the approximate density with the lowest reverse KL-divergence to the true posterior density, sampling one data point at a time \shortcite{blei2017variational,ganguly2021introduction}.
The objective function for this process can be formulated as:
\begin{align}
    \mathcal{D} &= \sum_{i=1}^{N}D_{\text{KL}}(q(z|x_{i}; \xi_{i})\;\|\;p(z|x_{i}; \theta)) = \sum_{i=1}^{N} \mathds{E}_{q}\bigg[\log \frac{q(z|x_{i}; \xi_{i})}{p(z|x_{i}; \theta)} \bigg],
    \label{equation:kld_1}
\end{align}
where $\mathds{E}_{q}\bigg[ \cdot \bigg] = \mathds{E}_{q(z|x_{i}; \xi_{i})}\bigg[ \cdot \bigg]$ and the output of this optimization process is the set of variational parameters that characterize the best approximation to the true posterior density.
Thus, for each local variational parameter $\xi_{i}$, inference amounts to solving the following optimization problem,
\begin{equation}
\label{equation:kld_opt}
    q^{*}(z|x_{i}; \xi_{i}) = \argmin_{q(z|x_{i}; \xi_{i}) \in \mathcal{Q}} D_{\text{KL}}(q(z|x_{i}; \xi_{i})\;\|\;p(z|x_{i}; \theta))
\end{equation}

The forward KL-divergence, $D_{\text{KL}}\big(p(z|x_{i}; \theta)\;\|\;q(z|x_{i}; \xi_{i})\big)$, can also be used as a measure in the objective function in Equation \ref{equation:kld_1} as opposed to the defined reverse KL-divergence, given its asymmetric nature.
However, in the case of VI, the forward KL-divergence cannot be computed in closed form as it requires taking expectations with respect to the unknown posterior.
For readers interested in understanding the foundational difference between forward and reverse KL-divergence, refer to \shortciteA{Murphy1991}.

Throughout this paper, we treat the global generative model parameter as a learnable parameter that is learned jointly with the variational parameters during the optimization process.
This is to ensure that the narrative of this paper is in line with the recent developments in the field of generative modeling, particularly relating to popular frameworks such as VAE \shortcite{kingma2013auto,rezende2015variational} and Generative Adversarial Networks (GANs) \shortcite{NIPS2014_5ca3e9b1}. 
While being two separate generative modeling frameworks, \shortciteA{jianlinsu} proves that GANs, like VAEs, are a special case of VI and proposes a unified framework between the two by re-formulating the VI objective.

\subsection{Evidence Lower Bound}

In addition to approximating the intractable posterior probability density for statistical inference over the latent variables, VI also enables efficient computation of a lower bound to the marginal likelihood or the evidence \shortcite{ganguly2021introduction}.
From the perspective of data modeling, a better fit to the observed data by a statistical model requires a better estimate of the evidence by that model.
Thus, it indicates that the chosen statistical model generating data points has a greater chance of being from the true data distribution.
Furthermore, this computation of the lower bound offers an alternative to the optimization problem defined in Equation \ref{equation:kld_opt} which is non-computable as it requires computing the evidence $\log p(x_{i})$ at each data point.

The KL-divergence objective function, for traditional VI, defined in Equation \ref{equation:kld_1} can be expanded as:
\begin{align}
    \mathcal{D} &= \sum_{i=1}^{N}D_{\text{KL}}(q(z|x_{i}; \xi_{i})\;\|\;p(z|x_{i}; \theta)), \nonumber\\
    &= \sum_{i=1}^{N} \mathds{E}_{q}\bigg[\log \frac{q(z|x_{i}; \xi_{i})}{p(z|x_{i}; \theta)} \bigg], \nonumber\\
    &= \sum_{i=1}^{N} \mathds{E}_{q}\bigg[ \log q(z|x_{i}; \xi_{i}) - \log p(z|x_{i};\theta)\bigg], \nonumber \\
    & = \sum_{i=1}^{N} \mathds{E}_{q}\bigg[ \log q(z|x_{i}; \xi_{i}) - \log \frac{p(x_{i}, z; \theta)}{p(x_{i}; \theta)}\bigg], \nonumber \\
    & = \sum_{i=1}^{N} \mathds{E}_{q}\bigg[ \log q(z|x_{i}; \xi_{i}) - \log p(x_{i}, z; \theta)\bigg] + \sum_{i=1}^{N} \mathds{E}_{q} \bigg[\log p(x_{i}; \theta)\bigg].
    \label{equation:elbo_1_1}
\end{align}
%
As the marginal likelihood is composed of a sum over the log marginal likelihoods of $N$ i.i.d. observations \shortcite{kingma2013auto}, i.e., 
\begin{equation*}
    \sum_{i}^{N}\log p(x_{i};\theta) = \log p(x_{1},..., x_{N}; \theta) = \log p(x; \theta),
\end{equation*}
 we re-write Equation \ref{equation:elbo_1_1} as:
\begin{align}
    \mathcal{D} &= \sum_{i=1}^{N} \mathds{E}_{q}\bigg[ \log q(z|x_{i}; \xi_{i}) - \log p(x_{i}, z; \theta)\bigg] + \log p(x; \theta), \nonumber \\
     -\mathcal{D} +  \log p(x; \theta) &= \sum_{i=1}^{N} \mathds{E}_{q}\bigg[\log p(x_{i}, z; \theta) - \log q(z|x_{i}; \xi_{i}) \bigg].
     \label{equation:elbo_2}
\end{align}
The sum of the negative KL-divergence and the log evidence in Equation \ref{equation:elbo_2} is referred to as the \textit{Evidence Lower Bound} (ELBO).
Equation \ref{equation:elbo_2} indicates that maximizing the ELBO with respect to $q$ is equivalent to minimizing the objective function defined in Equation \ref{equation:kld_1}.
In the case of traditional VI, we denote the ELBO by $\mathcal{L}(x; \xi_{1:N}, \theta)$ as follows:
\begin{align}
\label{equation:elbo_5}
     \mathcal{L}(x; \xi_{1:N}, \theta) &= \sum_{i=1}^{N} \mathds{E}_{q}\bigg[\log p(x_{i}, z; \theta) - \log q(z|x_{i}; \xi_{i}) \bigg] = \sum_{i=1}^{N}\mathcal{L}(x_{i}; \xi_{i}, \theta).
\end{align}
Furthermore, the ELBO, as evident from its name, is a lower bound estimate on the log marginal probability density of the data.
This can be derived using Jensen's inequality \shortcite{klarivcicsome} as:
\begin{align}
\label{equation:elbo_4}
\log p(x;\theta) &= \sum_{i=1}^{N}\log \int p(x_{i},z; \theta)\mathrm{d}z,\nonumber \\
&= \sum_{i=1}^{N}\log \int p(x_{i},z; \theta)\frac{q(z|x_{i}; \xi_{i})}{q(z|x_{i}; \xi_{i})}\mathrm{d}z, \nonumber \\
&= \sum_{i=1}^{N}\log \mathds{E}_{q}\biggl[\frac{p(x_{i},z; \theta)}{q(z|x_{i}; \xi_{i})}\biggr], \nonumber \\
&\geq \sum_{i=1}^{N}\mathds{E}_{q} \bigg[\log p(x_{i},z; \theta) - \log q(z|x_{i}; \xi_{i}) \bigg], \nonumber \\
\log p(x;\theta) &\geq \mathcal{L}(x; \xi_{1:N}, \theta).
\end{align}
An alternative way to show that $\mathcal{L}(x; \xi_{1:N}, \theta)$ is the lower bound of evidence is from Equations \ref{equation:elbo_2} and \ref{equation:elbo_5}:
\begin{align*}
    &\mathcal{L}(x; \xi_{1:N}, \theta) = \log p(x;\theta) - \mathcal{D}, \\
    &\log p(x;\theta) = \mathcal{L}(x; \xi_{1:N}, \theta) + \mathcal{D}.
\end{align*}
Since, the KL-divergence is non-negative, so $\log p(x;\theta) \geq \mathcal{L}(x; \xi_{1:N}, \theta)$.
This relationship establishes ELBO as a lower bound estimate of the incomplete log-likelihood of the data and, thus, at times, is used as a basis for selecting models to fit the data distribution.
\subsection{Mean Field Variational Family}

As computing the ELBO in Equation \ref{equation:elbo_5} requires taking expectations with respect to $q$, most applications using VI restrict the family of distributions, $\mathcal{Q}$, to be from the exponential family due to their conjugate nature leading to ease of computation \shortcite{wainwright_jordan_2007}.
An alternative way to ease this computation is to partition the elements of the latent vector $z$ into disjoint groups denoted by $z_{k}$ where $k=1,...,N$.
Thus, assuming the variational posterior to be factorized as:
\begin{equation}
\label{equation:mean_field}
    q(z|x_{i}; \xi_{i}) = \prod_{k=1}^{N}q_{k}(z_{k}|x_{i}; \xi_{i}).
\end{equation}
This factorized form of VI corresponds to a framework developed in physics called mean field theory \shortcite{parisi1988statistical} and is known as mean field VI \shortcite{opper_saad_2001}.
In the context of belief networks, the mean field theory was further developed by \shortciteA{BhattacharyyaK01}.
It is to be noted that each of these variational factors can take on any parametric form appropriate to the corresponding random variable \shortcite{blei2017variational}.
The ELBO is maximized with respect to each of these factors in Equation \ref{equation:mean_field} which on substitution into Equation \ref{equation:elbo_5} and denoting $q_{k}(z_{k}|x_{i}; \xi_{i})$ as $q_{k}$ for notational clarity, we obtain,
\begin{align}
\label{equation:mean_field_1}
    \mathcal{L}(x; \xi_{1:N}, \theta)&= \sum_{i=1}^{N} \mathds{E}_{q}\bigg[\log p(x_{i}, z; \theta) - \log q(z|x_{i}; \xi_{i}) \bigg], \nonumber \\
    &= \sum_{i=1}^{N} \int \prod_{k}q_{k}\bigg[\log p(x_{i}, z; \theta) - \log \prod_{k}q_{k}\bigg]\mathrm{d}z, \nonumber \\
    &= \sum_{i=1}^{N} \bigg\{\int q_{j}\bigg[\int \log p(x_{i}, z; \theta)\prod_{k\neq j}q_{k}\mathrm{d}z_{k}\bigg]\mathrm{d}z_{j} - \int q_{j}\log q_{j}\mathrm{d}z_{j}\bigg\} \nonumber, \\
    &= \sum_{i=1}^{N} \bigg\{\int q_{j}\mathds{E}_{k \neq j}\bigg[\log p(x_{i}, z; \theta)\bigg]\mathrm{d}z_{j} - \int q_{j}\log q_{j}\mathrm{d}z_{j} - \mathcal{H}_{k \neq j}\bigg\},
\end{align}
where $\mathcal{H}_{k \neq j}$ and $\mathds{E}_{k \neq j}\bigg[\cdot \cdot \cdot \bigg]$ denote the entropy and the expectation with respect to probability densities over all latent variables $z_{k}$ for $k \neq j$. The full derivation for Equation \ref{equation:mean_field_1} is shown in Appendix \ref{appendix:appendix_A}.

The ELBO in Equation \ref{equation:mean_field_1} is maximized repeatedly with respect to each of the factors, $q_{j}$, while keeping the remaining factors, $q_{k \neq j}$ constant.
Convergence is guaranteed because the bound is convex with respect to each of the factors $q_{i}$ \shortcite{boyd_vandenberghe_2004}.

An extension to the mean field VI formulation is structured VI \shortcite{saul_mfvi,barber_nips99}, which adds dependencies between the variables leading to a better approximation of the posterior probability density.
There is, however, a trade-off as introducing these dependencies may make the variational optimization problem difficult to solve. 

\subsection{Coordinate Ascent Optimization}

As for the optimization process, the coordinate ascent VI algorithm (CAVI) has been a popular choice for solving the traditional VI problem \shortcite{10.5555/1162264,hoffmanstochastic,blei2017variational,plummer2020dynamics} as it complements the mean field VI optimization process. 
The coordinate ascent algorithm can look like the EM algorithm\footnote{Interested readers are requested to read Chapter 11.4 of \shortciteA{Murphy1991}.} where the “E step” computes approximate conditionals of local latent variables and the “M step” computes a conditional of the global latent variables \shortcite{blei2017variational}.
Similar to mean field VI, this optimization process works by repeatedly updating each random variable’s variational parameters based on the variational parameters of the variables in its Markov blanket\footnote{The Markov Blanket of a target variable is a minimal set of variables that the target variable is conditioned on while all other remaining variables in the model are probabilistically independent of the target variable \shortcite{TsamardinosAS03}.} \shortcite{winn_bishop}, and re-estimating the convergence of the ELBO (described in Algorithm \ref{algoritm:cavi}).
CAVI goes uphill on the ELBO of Equation \ref{equation:elbo_5}, eventually finding a local optimum \shortcite{blei2017variational}.
\shortciteA{ganguly2021introduction} illustrate an example to approximate a mixture of Gaussians using coordinate ascent and mean field VI.

\vspace*{.2cm}
\begin{algorithm}[H]
\SetAlgoLined
\DontPrintSemicolon
\KwIn{Data \(x_{1:N}\)}
\KwOut{Variational parameters \(\xi_{1:N}\); generative model parameter $\theta$}
$ \theta, \xi_{1:N} \gets$ random initialization \\
\While{$\mathcal{L}(x; \xi_{1:N}, \theta)$ has not converged}{
    \For{$i \in  {1, ..., N}$}{
        $\displaystyle$ \\
        $\displaystyle \xi_{i} \gets \argmax_{\xi_{i}} \mathcal{L}(\xi_{i}, \theta; x_{i})$\\
        }
    Compute $\mathcal{L}(x; \xi_{1:N}, \theta)$
    $\displaystyle$ \\
    $\displaystyle \theta \gets \argmax_{\theta} \sum_{i=1}^N\mathcal{L}(x_{i}; \xi_{i}, \theta)$ \\
    }
\Return{$\xi_{1:N}$; $\theta$}
\caption{CAVI for the traditional VI optimization process}
\label{algoritm:cavi}
\end{algorithm}
\vspace*{.2cm}

Though this optimization process results in a closed-form solution for the optimal variational parameters, it is inefficient for large data sets as it requires a complete pass through the entire data set, sampling one data point at a time, at each iteration.
Generally, the ELBO is a non-convex objective function and CAVI guarantees convergence to a local optimum and is sensitive to initialization \shortcite{blei2017variational}.
Furthermore, in combination with the mean field approximations, CAVI may lead to sub-optimal convergence as the former explicitly ignores correlations between variables, thereby making the optimization problem more non-convex \shortcite{wainwright_jordan_2007}.

\section{Stochastic VI}
\label{section:svi}

In recent years, with the advent of big data, scalability, and efficiency have become the primary requirements for modern machine learning algorithms.
In the field of VI, one notable development has been in the form of \textit{stochastic variational inference} (SVI) \shortcite{hoffmanstochastic} which combines natural gradients \shortcite{amari1998natural} and stochastic optimization \shortcite{Robbins_1951} to tackle the scalability issue of the traditional VI algorithm.

\subsection{Stochastic Optimization}

In contrast to CAVI, which updates the variational parameters one data point at a time, SVI uses stochastic optimization \shortcite{Robbins_1951}, following noisy estimates of the gradient of the ELBO, on a sub-sample of the data and updates the parameters based on that sub-sample.

We can re-construct the ELBO, formulated in Equation \ref{equation:elbo_5}, an estimator of the full data set, based on sets of mini-batches, $x^{M}$ as:
\begin{equation}
   \mathcal{L}(x; \xi_{1:N}, \theta) \simeq \frac{N}{M}\sum_{i=1}^{M}\mathcal{L}(x_{i}; \xi_{i}, \theta) =\frac{N}{M}\mathcal{\hat{L}}(x^{M}; \xi^{M}, \theta),
    \label{equation:svi}
\end{equation}
where $M$ is the randomly drawn sub-sample of data from $N$ data points, $x^{M}$ represents a random mini-batch of the data set, and $\sum_{i=1}^{M}\mathcal{L}(x_{i}; \xi_{i}, \theta)$ is an estimate of the ELBO based on those $M$ samples, which we denote as $\mathcal{\hat{L}}(x^{M}; \xi^{M}, \theta)$.

The methodology of SVI is to get a stochastic estimate of the ELBO based on a set of $M$ examples at each iteration (with or without replacement).
This allows us to take derivatives $\nabla_{\xi^{M}, \theta}\mathcal{\hat{L}}(x^{M}; \xi^{M}, \theta)$ and update the local variational parameters based on the $M$ samples as well as the global parameter, $\theta$, using stochastic gradient ascent.
We repeat this process until the ELBO converges.
Computational savings in SVI are obtained only for $M \ll N$ \shortcite{zhang_vi_2019}.

\subsection{Natural Gradients}
\label{section:natural_grad}

\shortciteA{hoffmanstochastic} proposed the idea of using \textit{natural} gradients as opposed to using standard gradients for SVI to capture the information geometry of the parameter space for probability densities. 
Natural gradients adjust the direction of the traditional gradient by the use of a Riemannian metric.
The classical gradient ascent method for a function $f(\xi)$ tries to reach the function's maxima by taking steps of size $\rho$ in the direction of the steepest ascent for the gradient (when it exists) \shortcite{hoffmanstochastic}.
This is formulated as:
\begin{equation*}
    \xi^{t+1} = \xi^{t} + \rho \nabla_{\xi}f(\xi^{t}),
\end{equation*}
where the gradient $\nabla_{\xi}f(\xi)$ points in the same direction as the solution to
\begin{equation*}
    \argmax_{\Delta\xi}f(\xi + \Delta\xi) \quad\text{subject to}\quad \|\Delta\xi\|^2 < \epsilon^2 \quad\text{and}\quad \epsilon \rightarrow 0.
\end{equation*}
During optimization, satisfying the condition above enables a movement away from $\xi$ in the direction of the gradient. 
It is clear that in classical gradient ascent, the gradient direction depends on the Euclidean distance metric associated with the space where $\xi$ resides.
However, when optimizing an objective involving parameterized probability density functions, the Euclidean distance between two parameter vectors $\xi$ and $\xi + \Delta\xi$ is often a poor measure of the dissimilarity of the probability densities $q(z; \xi)$ and $q(z; \xi + \Delta\xi)$ \shortcite{hoffmanstochastic}.
This is because the Euclidean metric fails to offer a meaningful explanation of distance in spaces where the local distance is not defined by the L2 norm.

Natural gradient corrects this issue by redefining the criterion for the gradient's motion in the direction of the steepest ascent as:
\begin{equation*}
    \argmax_{\Delta\xi}f(\xi + \Delta\xi) \quad\text{subject to}\quad  D^{\text{sym}}_{\text{KL}}(\xi, \xi+\Delta\xi)< \epsilon^2 \quad\text{and}\quad \epsilon \rightarrow 0,
\end{equation*}
where $D^{\text{sym}}_{\text{KL}}(\xi, \xi+\Delta\xi)$ is the symmetrized KL-divergence which is defined as:
\begin{equation}
    D^{\text{sym}}_{\text{KL}}(\xi,\xi+\Delta\xi) = \mathds{E}_{q(z; \xi)}\bigg[\log\frac{q(z; \xi)}{q(z; \xi+\Delta\xi)}\bigg] + \mathds{E}_{q(z; \xi + \Delta\xi)}\biggl[\log\frac{q(z; \xi+\Delta\xi)}{q(z; \xi)} \bigg].
\end{equation}

While the Euclidean gradient points in the direction of steepest ascent in an Euclidean space, the natural gradient points in the direction of steepest ascent in the Riemannian space -- a space where local distance is defined by KL-divergence rather than the L2 norm \shortcite{hoffmanstochastic}.
In higher dimensions, using natural gradients, a movement of the same distance in different directions amounts to an equal change in the symmetrized KL-divergence \shortcite{blei2017variational}.
\shortciteA{do_carmo_1993} introduced a Riemannian metric, $I(\xi)$, which defines the distance between $\xi$ and a nearby vector $\xi+\Delta\xi$ as:
\begin{equation*}
    \Delta\xi^{T}I(\xi)\Delta\xi \approx D^{\text{sym}}_{\text{KL}}(\xi,\xi+\Delta\xi),
\end{equation*}
where $I(\xi)$ is the Fisher information matrix of $q(z; \xi)$. 
The full derivation is shown in Appendix \ref{appendix:appendix_B}.
\shortciteA{amari_1982} showed that natural gradients can be obtained by pre-multiplying the gradients with the inverse Fisher information matrix as:
\begin{equation*}
    \Bar{\nabla}_{\xi}f(\xi) \triangleq [I(\xi)]^{-1}\nabla_{\xi}f(\xi).
\end{equation*}
where $\Bar{\nabla}$ and $\nabla$ denote natural and stochastic gradients respectively.

The Fisher information matrix is essential to compute the Cram{\'e}r-Rao lower bound for the performance analysis of an unbiased estimator --- a minimum variance estimator for a parameter \shortcite{merberg2008course,yang1997efficiency}.
In VI, for a high dimensional parameter space, studying the covariance matrix for the variational estimator provides an estimate for its unbiasedness.
The underlying high dimensional posterior structure might be rich, and the covariance matrix for the variational parameters captures the uncertainty of the KL-divergence being locked onto one of its many local modes.
Additionally, it captures the sensitivity of the estimated posterior density with respect to small variations in the variational parameters \shortcite{knollmuller2019metric}.
For each of the variational parameters, $\xi_{i}$, to be unbiased estimators of the true parameters, they must satisfy the Cram{\'e}r-Rao bound as:
\begin{equation}
    \text{cov}(\xi) \geq [I(\xi)]^{-1}.
\end{equation}

For the ELBO formulation in Equation \ref{equation:elbo_5}, the Fisher information matrix for a variational parameter $\xi_{i}$ is computed as:
\begin{equation*}
    I(\xi_{i}) := \mathds{E}_{q}\bigg[ \nabla_{\xi_{i}} \log q(z|x_{i}; \xi_{i}) \nabla_{\xi_{i}} \log q(z|x_{i}; \xi_{i})^T\bigg],
\end{equation*}
and for the generative model parameter $\theta$, it is computed as:
\begin{equation*}
    I(\theta) := \mathds{E}_{p(x_{i}, z; \theta)}\bigg[ \nabla_{\theta} \log p(x_{i}, z; \theta) \nabla_{\theta} \log p(x_{i}, z; \theta)^T\bigg].
\end{equation*}
For a given step size $\rho > 0$, the natural gradient updates for the parameters at a time step $t+1$ is given by:
\begin{align*}
    \xi_{i}^{t+1} &= \xi_{i}^{t} + \rho [I(\xi_{i}^t)]^{-1}\nabla_{\xi_{i}}\mathcal{L}(x_{i}; \xi_{i}^t, \theta^{t}),
\end{align*}
and
\begin{align*}
    \theta^{t+1} &= \theta^{t} + \rho [I(\theta^t)]^{-1}\nabla_{\theta}\mathcal{L}(x_{i}; \xi_{i}^t, \theta^{t}).
\end{align*}
Additionally, the Fisher information matrix is a measure of the curvature for a probability density function as it is equal to the expected Hessian for that density function \shortcite{martens_grad_hess} (see Appendix \ref{appendix:appendix_C}).
This property is useful in problems wherein the Fisher information matrix is infeasible to compute, store, or invert.
In such cases, simply computing the second moment of the derivatives is equivalent to approximating the Fisher information matrix.

\afterpage{
\begin{algorithm}
 \caption{The SVI optimization process based on \shortciteA{hoffmanstochastic}}
 \label{algoritm:svi}
  \KwIn{Data \(x_{1:N}\)}
  \KwOut{Variational parameter \(\xi_{1:N}\); generative model parameter $\theta$}
  $\xi_{1:N},  \theta \gets$ random initialization\\
  \While{$\mathcal{L}(x_{i}; \xi_{1:N}, \theta)$ has not converged}{
  \Repeat{all $N$ data points are seen at least once}{
    $\mathcal{M} \gets \{x^{M} | x^{M} \sim x_{1:N}$, $|x^{M}| = M\}$ \\
    \For{$x^{M} \in \mathcal{M}$}{
    $\displaystyle$ \\
    \For{$x_{j} \in x^{M}$}
    {$\displaystyle$ \\
      Compute $\mathcal{L}(x_{j}; \xi_{j}, \theta)$\\
      Compute $\nabla_{\xi_{j}}\mathcal{L}(x_{j}; \xi_{j}, \theta); \nabla_{\theta}\mathcal{L}(x_{j}; \xi_{j}, \theta)$ \\
    }
    $\displaystyle \hat{\nabla}_{\xi^M}\mathcal{L}(x^{M}; \xi^{M}, \theta) = \frac{1}{M}\sum\limits_{j} \nabla_{\xi_{j}}\mathcal{L}(x_{j}; \xi_{j}, \theta)$ \\
    $\displaystyle \hat{\nabla}_{\theta}\mathcal{L}(x^{M}; \xi^{M}, \theta) = \frac{1}{M}\sum\limits_{j} \nabla_{\theta}\mathcal{L}(x_{j}; \xi_{j}, \theta)$ \\
    \For{$x_j \in x^{M}$}
    {$\displaystyle$ \\
      $\displaystyle \Bar{\nabla}_{\xi_{j}}\mathcal{L}(x^{M}; \xi^{M}, \theta) \triangleq [I(\xi_{j})]^{-1}\hat{\nabla}_{\xi^M}\mathcal{L}(x^{M}; \xi^{M}, \theta)$  \\
      $\displaystyle \xi_{j} \gets$ Update parameters using $\displaystyle \Bar{\nabla}_{\xi^{M}}\mathcal{L}(x^{M}; \xi^{M}, \theta)$\\
    }
    $\displaystyle$ \\
    $\Bar{\nabla}_{\theta}\mathcal{L}(x^{M}; \xi^{M}, \theta) \triangleq [I(\theta)]^{-1}\hat{\nabla}_{\theta}\mathcal{L}(x^{M}; \xi^{M}, \theta)$ \\
    $\displaystyle \theta \gets$ Update parameters using $\Bar{\nabla}_{\theta}\mathcal{L}(x^{M}; \xi^{M}, \theta)$\\
    $\displaystyle \mathcal{\hat{L}}(x^{M}; \xi^{M}, \theta) = \sum_{j=1}^{M}\mathcal{L}(x_{j}; \xi_{j}, \theta)$ \\
    }
    }
    $\displaystyle \mathcal{L}(x; \xi_{1:N}, \theta) \simeq \frac{N}{M}\mathcal{\hat{L}}(x^{M}; \xi^{M}, \theta)$
}
\Return{$\xi_{1:N}$; $\theta$}
\end{algorithm}
%
}
The full SVI algorithm using mini-batches and natural gradients is described in Algorithm \ref{algoritm:svi}.
This SVI methodology has aided in significant advancements in VI, such as gamma processes \shortcite{knowles2015stochastic} and more specifically in the development of the VAE \shortcite{kingma2013auto,RezendeMW14} and its different variants.

\subsection{Faster Convergence in SVI}
\label{section:faster}

The speed of convergence for the SVI optimization process depends on the variance of the gradient estimates.
A lower variance ensures minimum gradient noise allowing for larger learning rates, leading to faster convergence.
One approach to reducing the variance is to increase the mini-batch size, which leads to lower gradient noise as suggested by the law of large numbers \shortcite{svi_hmm}.
Another alternative is to use non-uniform sampling, such as importance sampling, to select mini-batches with lower gradient noise.
Researchers have proposed different variants of importance sampling \shortcite{csiba_imp,gopalan,parisi1988statistical,zhaoa15_imp} for this purpose. 
Although effective, the computational complexity of the sampling mechanism, however, relates to the dimensionality of model parameters \shortcite{fu_cpsg}.

Increasing the mini-batch size might not always be plausible owing to hardware memory constraints.
Recent trends in deep learning have shown that an increase in the speed of the training procedure can also be achieved by adjusting the learning rate while keeping the mini-batch size fixed.
The idea is to let the empirical gradient variance guide the adaptation of the learning rate which is inversely proportional to the gradient noise in each iteration \shortcite{zhang_vi_2019}.
Gradually adapting the learning rate guarantees that every point in the parameter space can be reached, while the gradient noise decreases sufficiently fast to ensure convergence \shortcite{Robbins_1951}.
Several optimization techniques such as Adam \shortcite{Kingma2015AdamAM}, AdaGrad
\shortcite{Duchi2010AdaptiveSM}, AdaDelta \shortcite{Zeiler2012ADADELTAAA} and  RMSProp \shortcite{tieleman2012lecture}, which make use of this idea, have been developed.

Other than increasing the mini-batch size or adapting the learning rate, variance reduction can be achieved using a control variate \shortcite{neural_cv}, a stochastic term, which when added to the stochastic gradient reduces the variance while keeping its expected value intact \shortcite{BOYLE1977323}.
Using control variates for variance reduction is common in Monte Carlo simulation and stochastic optimization \shortcite{zhang_vi_2019,ross_sim,wang_var_opt}.

\section{Black Box VI}
\label{section:bbvi}

The traditional VI process performs an initial analytical derivation for the ELBO before optimization which requires time and mathematical expertise.
Thus, this makes it limited for use with only conditionally conjugate exponential families \shortcite{hoffmanstochastic,zhang2016structured}.
For this purpose, \shortciteA{ranganath2014black} introduced the Black Box VI (BBVI) methodology that removes the need for analytical computation of the ELBO, relaxing this limitation.
Based on the SVI optimization process, BBVI computes the gradient from Monte Carlo samples generated from the variational probability density.

The gradient estimate for the ELBO formulated in Equation \ref{equation:elbo_5} at a data point $x_{i}$ can be written as:
\begin{equation}
\label{equation:elbo_grad}
    \nabla_{\xi_{i}}\mathcal{L}(x_{i}; \xi_{i}, \theta) =  \nabla_{\xi_{i}}\Bigg[\mathds{E}_{q}\bigg[\log p(x_{i}, z; \theta) - \log q(z|x_{i}; \xi_{i})\bigg] \Bigg].
\end{equation}
Drawing parallels from reinforcement learning, the variational probability density represents the policy as it is used to generate samples.
The reward is the ELBO maximizing which guides the iterative learning process for the optimal variational parameters.
An episode is formed by initializing the variational parameters and thus the policy, followed by drawing samples from the variational distribution, subsequently computing the ELBO or the reward, and updating the policy parameters. 
Thus, the gradient estimate of the ELBO as defined in Equation \ref{equation:elbo_grad} is similar to computing the derivative of the expected reward while following a policy in a reinforcement learning setting.
However, computing the gradient of the objective containing an expectation is non-trivial.
For this purpose, reinforcement learning problems make use of the policy gradient theorem \shortcite{Sutton00policygradient}, which states that the derivative of the expected reward is the expectation of the product of the reward and gradient of the log of the policy.
Thus, using the policy gradient theorem \shortcite{Sutton00policygradient}, Equation \ref{equation:elbo_grad} can be re-written as:
\begin{equation*}
    \nabla_{\xi_{i}}\mathcal{L}(x_{i}; \xi_{i}, \theta) = \mathds{E}_{q}\Bigg[\nabla_{\xi_{i}}\log q(z|x_{i}; \xi_{i})\bigg[\log p(x_{i}, z; \theta) - \log q(z|x_{i}; \xi_{i})\bigg]\Bigg].
\end{equation*}
The gradient $\nabla_{\xi_{i}}\mathcal{L}(x_{i}; \xi_{i}, \theta)$ involving expectation with respect to $q(z|x_{i}; \xi_{i})$ can be approximated by drawing $K$ independent samples, $z_{k}$, from the variational distribution and then computing the average of the function evaluated at these samples \shortcite{mohamed2020monte} as:
\begin{equation}
    \nabla_{\xi_{i}}\mathcal{L}(x_{i}; \xi_{i}, \theta) \approx \frac{1}{K} \sum_{k=1}^K \bigg[ \log p(x_{i}, z_{k}; \theta) - \log q(z_{k}|x_{i}; \xi_{i})\bigg] \nabla_{\xi_{i}}\log q(z_{k}|x_{i}; \xi_{i}), \label{equation:bbvi}
\end{equation}
where $z_{k} \sim q(z|x_{i}; \xi_{i})$ and $\nabla_{\xi_{i}}\log q(z_{k}|x_{i}; \xi_{i})$ is known as the score function \shortcite{cox_hinkley_1994}. It is to be noted that the score function and sampling algorithms depend only on the variational distribution, not the underlying model.
BBVI thus enables the practitioner to obtain an unbiased gradient estimator by sampling without having to derive the gradient of the ELBO explicitly \shortcite{zhang_vi_2019}.

However, the variance of the gradient estimator under the Monte Carlo estimate in Equation \ref{equation:bbvi} can be too large to be useful \shortcite{ranganath2014black}.
Unlike SVI, where sub-sampling from a finite set of data points leads to noisy gradient estimates, in the case of BBVI, it is the possible oversampling of the random variables that contributes to high noise in the gradients.
Researchers have developed several variance reduction techniques for BBVI, such as the combination of Rao-Blackwellization and control variates \shortcite{ranganath2014black}, local expectation gradients \shortcite{Titsias_bbvi}, and overdispersed importance sampling \shortcite{Ruiz2016OverdispersedBV} but most notably, the reparameterization trick, introduced by \shortciteA{kingma2013auto} (discussed in Section \ref{section:reparam}), is often used as it enables lower variance gradient estimates than the rest.

BBVI and its extensions have been one of the most significant developments of modern approximate inference techniques making amortized inference feasible in solving several deep learning problems \shortcite{set-transformer-dac2019,dac2019,dac2021,darm2022}.

\subsection{The Reparameterization Trick}
\label{section:reparam}

As established in Section \ref{section:faster} it is necessary to maintain a low variance for the stochastic gradient estimates to ensure faster convergence.
Both \shortciteA{ranganath2014black} and \shortciteA{kingma2013auto} state that the Monte Carlo gradient estimates in BBVI (Equation \ref{equation:bbvi}) exhibit high variance.
For this purpose, \shortciteA{kingma2013auto} introduced a more practical gradient estimator for the lower bound in the form of a reparameterization trick.
For a chosen approximate posterior $q(z|x_{i}; \xi_{i})$, the trick allows a random variable $z_{i}$ to be a differentiable transformation $g_{\phi}(\epsilon, x_{i})$ of a noise variable $\epsilon$, such that,
\begin{align*}
    z_{i} &= g_{\phi}(\epsilon, x_{i}), \\
    \epsilon &\sim p(\epsilon).
\end{align*}
Given a function $f(z)$, Monte Carlo estimates of expectations of it with respect to $q(z|x_{i}; \xi_{i})$ can be formed as follows:
\begin{equation*}
    \mathds{E}_{q} \bigg[f(z)\bigg] = \mathds{E}_{p(\epsilon)} \bigg[ f(g_{\phi}(\epsilon, x_{i})) \bigg] \simeq \frac{1}{K} \sum_{k=1}^{K} f(g_{\phi}(\epsilon_{k}, x_{i})) \quad\text{where}\quad \epsilon_{k} \sim p(\epsilon).
\end{equation*}
\shortciteA{kingma2013auto} show that applying this to the ELBO for VI in Equation \ref{equation:elbo_5}, yields the stochastic estimator $\mathcal{\Tilde{L}}(x; \xi_{1:N}, \theta) \simeq \mathcal{L}(x; \xi_{1:N}, \theta)$:
\begin{align}
\label{equation:elbo_sgvb}
    \mathcal{\Tilde{L}}(x; \xi_{1:N}, \theta) &= \sum_{i=1}^{N} \Bigg[ \frac{1}{K} \sum_{k=1}^{K}\bigg[\log p(x_{i}, z_{(i,k)}; \theta) - \log q(z_{(i,k)}|x_{i}; \xi_{i}) \bigg] \Bigg], \nonumber \\
    &= \sum_{i=1}^{N} \mathcal{\Tilde{L}}(x_{i}; \xi_{i}, \theta), 
\end{align}
where $z_{(i,k)} = g_{\phi}(\epsilon_{(i, k)}, x_{i})$, $\epsilon_{k} \sim p(\epsilon)$ and $\mathcal{\Tilde{L}}(x_{i}; \xi_{i}, \theta)$ is the stochastic estimator of the ELBO at a data point $x_{i}$.
The gradient $\nabla_{\xi_{i}} \mathcal{\Tilde{L}}(x_{i}; \xi_{i}, \theta)$ for the estimator in Equation \ref{equation:elbo_sgvb} can thus be written as:
\begin{equation}
\label{equation:elbo_grad_sgvb}
    \nabla_{\xi_{i}} \mathcal{\Tilde{L}}(x_{i}; \xi_{i}, \theta) = \frac{1}{K}\sum_{k=1}^{k} \nabla_{\xi_{i}} \Bigg[ \log p(x_{i}, z_{(i,k); \theta}) - \log q(z_{(i,k)}|x_{i}; \xi_{i}) \Bigg].
\end{equation}
Comparing Equation \ref{equation:elbo_grad_sgvb} with the policy gradient formulation for BBVI in Equation \ref{equation:bbvi}, we see that the gradient of the log joint distribution is a part of the expectation.
The advantage of taking the gradient of the log joint is that this term is more informed about the direction of the maximum posterior mode \shortcite{zhang_vi_2019}.
This information also contributes to lower variance for the gradient estimates when compared to the policy gradient estimates.
However, the ELBO in Equation \ref{equation:elbo_grad_sgvb} suffers from injected noise due to the use of Monte Carlo estimation for the lower bound.
This noise can further be reduced by the use of control variates \shortcite{foti} or Quasi-Monte Carlo methods \shortcite{buchholz2018quasi}.
Additionally, like BBVI the reparameterization trick allows us to derive the ELBO without having to compute analytic expectations.
This reparameterization trick is also the basis of VAEs \shortcite{kingma2013auto,Robbins_1951}.

Although it promises a lower variance for the gradient estimates, the reparameterization trick, unlike the policy gradient scheme for BBVI, does not trivially extend to discrete distributions.
In order for the trick to be applied to discrete distributions, further approximations for the variational posterior are required \shortcite<see>{gu17,MaddisonMT17,NalisnickS17}.

\section{Amortized VI}
\label{section:avi}
The traditional VI optimization problem, as described in Section \ref{section:vi}, maximizes the ELBO with respect to the variational parameters for each data point, followed by computing the optimal global generative model parameters as:
\begin{equation*}
    \argmax_{\theta}\sum_{i=1}^{N}\argmax_{\xi_{i}}\mathcal{L}(x_{i}; \xi_{i}, \theta).
\end{equation*}
This repetitive process introduces a new set of variational parameters for every observation, allowing these parameters to grow, at least, linearly with the observations.
During this optimization process, each observation is processed independently of others, making the process \textit{memoryless}, thereby guaranteeing that inference using one observation will not interfere with another \shortcite{gershman2014amortized}.
This further implies that there is no mechanism to re-use the knowledge from previous inferences on newer ones, and as such inferring on the same observation twice requires the same amount of computation which is equivalent to inferring two separate ones  \shortcite{gershman2014amortized}. 
When the number of observations is large, it can also lead to extensive computational inefficiency since there is no memory trace of inferences from previous data points. 
It might, therefore, be helpful to keep a memory trace of the past inferences (\textit{memoizing}), although at a higher cost, to solve this scalability issue.
However, it may be inaccurate to re-use a stored inference without modification as newer observations might be related or modifications to previous ones.

\afterpage{
\begin{longtable}{ll}
\hline
Method & Properties \\ 
\hline
\endfirsthead
\caption{Comparison of different VI methods (continued on the next page).} 
\endfoot
\endlastfoot
Traditional VI & \textbf{Methodology}: \\ 
  & $\bullet$ Analytical approximation of the posterior probability density for \\
  & statistical inference over latent variables.\\ 
  & $\bullet$ Formulates statistical inference as an optimization problem using a \\
  & suitable divergence measure.\\ 
  & $\bullet$ Introduces a local variational parameter for every observation.\\
  & $\bullet$ Uses coordinate ascent to optimize the variational parameters for \\
  & each observation iteratively.\\ 
  & \\ 
  & \textbf{Advantages}:\\ 
  & $\bullet$ Use the ELBO to tractably compute the evidence to encourage the \\
  & chosen statistical model to fit the data better.\\ 
  & \\ 
  & \textbf{Limitations}:\\ 
  & $\bullet$ Inefficient in scaling to large datasets.\\ 
  & $\bullet$ Coordinate ascent may encourage convergence to a local optimum. \\
  & $\bullet$ Requires analytical derivation of the ELBO. \\
\hline
SVI & \textbf{Methodology}: \\ 
  & $\bullet$ Uses gradient-based optimization to update the local variational \\
  & parameters. \\ 
  & $\bullet$ Optimization is based on mini-batches of data rather than iterating \\
  & over every observation.\\ 
  & $\bullet$ Can be combined with natural gradients to capture the \\
  & dissimilarities of probability densities efficiently.\\ 
  & \\ 
  & \\ 
  & \textbf{Advantages}: \\ 
  & $\bullet$ Variance reduction of the gradients can be achieved by either \\
  & increasing the mini-batch size or adjusting the learning rate during \\
  & training or using control variates.\\ 
  & $\bullet$ Fast convergence.\\ 
  & $\bullet$ Scalable to large datasets.\\ 
  & \\ 
  & \textbf{Limitations}:\\ 
  & $\bullet$ Still requires analytical derivation of the ELBO. \\
\hline
BBVI & \textbf{Methodology}: \\ 
  & $\bullet$ Uses gradient-based optimization to update the local variational \\
  & parameters. \\ 
  & $\bullet$ Optimization is based on mini-batches of data rather than iterating \\
  & over every observation.\\ 
  & $\bullet$ Uses the reparameterization trick to maintain a low variance for the \\
  & stochastic gradient estimates for the ELBO.\\ 
  & \\ 
  & \textbf{Advantages}:\\ 
  & $\bullet$ Omits the requirement to derive the ELBO analytically. \\ 
  & $\bullet$ Fast convergence.\\ 
  & $\bullet$ Scalable to large datasets.\\ 
  & \\ 
  & \textbf{Limitations}:\\ 
  & $\bullet$ The reparameterization trick does not extend to discrete distributions.\\
\hline
Amortized VI & \textbf{Methodology}: \\ 
  & $\bullet$ Amortizes the inference by the use of a stochastic function, such as a\\
  & neural network, to map the observed variables to the latent variables.\\ 
  & $\bullet$ Uses the BBVI methodology for ELBO optimization.\\ 
  & \\ 
  & \textbf{Advantages}:\\ 
  & $\bullet$ Flexible memoized re-use of past inferences to compute inference on \\
  & newer observations.\\ 
  & $\bullet$ Omits the requirement to derive the ELBO analytically.\\ 
  & $\bullet$ Fast convergence.\\ 
  & $\bullet$ Scalable to large datasets.\\ 
  & \\ 
  & \textbf{Limitations}:\\ 
  & $\bullet$ The use of a stochastic function to amortize the inference leads to \\
  & inconsistent representation learning.\\ 
  & $\bullet$ Sub-optimal inference arising largely due to a coding efficiency gap \\
  & known as amortization gap.\\ 
  & $\bullet$ Generalization gap depends on the capacity of chosen neural network \\
  & as the stochastic function. \\
\hline
\caption{Comparison of different VI methods.} 
\label{tab:compare_VI_methods}
\end{longtable}
}

In the case of VI, these issues are solved by ``amortizing" the optimization process, where instead of optimizing for each data point independently, the optimization cost is spread out across multiple instances, reducing the overall computational burden \shortcite{Amos2022TutorialOA}.
For this purpose, amortized VI makes use of a stochastic function, which maps the observed variable to the latent variable belonging to the variational posterior density, the parameters of which are learned during the optimization process.
Therefore, instead of having separate parameters for each observation, the estimated function can infer latent variables even for new data points without re-running the optimization process all over again on the new data points.
This process allows for computational efficiency and flexible memoized re-use \shortcite{michie1968memo} of relevant information from past inferences on previously unseen data.
Table \ref{tab:compare_VI_methods} compares the methodology, advantages, and limitations of traditional VI, SVI, BBVI, and amortized VI.

With recent advances in deep learning, researchers have extensively used neural networks in the form of this stochastic function to estimate the parameters of the posterior probability density.
Neural networks are powerful frameworks that allow for efficient amortization of inference. 
Additionally, the development of GPU-assisted neural network training has also led to the usage of complex neural network architectures with amortized VI \shortcite<e.g.,>{radford2015unsupervised,karras2019style,chen2016infogan,pidhorskyi2020adversarial}, allowing extraction of information from high-dimensional data without human supervision \shortcite{simonyan2013deep}.

\begin{figure}
    \centering
    \includegraphics[clip, width=0.4\columnwidth]{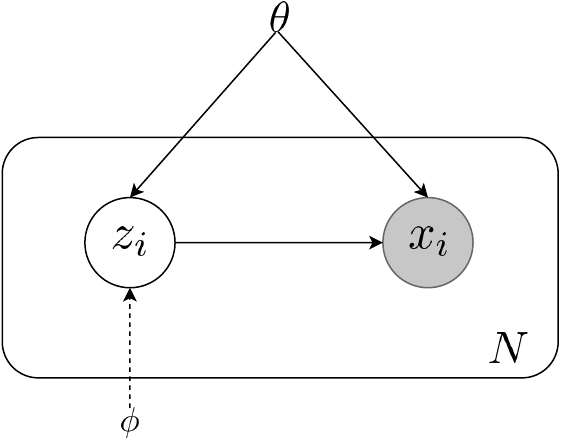}
    \caption{ Illustration of the directed graphical model in the case of amortized VI with \(N\) observed data points. The global and the amortized variational parameters are represented by $\theta$ and $\phi$ respectively.
    }
    \label{fig:pgm_avi}
\end{figure}

While a local variational parameter, $\xi_{i}$, is introduced for every observation, $x_{i}$, in traditional VI, as shown in Figure \ref{fig:pgm}, in case of amortized VI, the variational parameters, $\phi$, are globally shared by all the observations, illustrated by the graphical model in Figure \ref{fig:pgm_avi}.
Thus, the ELBO defined for the traditional VI optimization problem, established in Equation \ref{equation:elbo_5}, can be modified for amortized VI as:
\begin{equation}
\label{equation:elbo_avi}
\mathcal{L}(x; \phi, \theta) = \sum_{i=1}^{N} \mathds{E}_{q_{\phi}}\bigg[\log p(x_{i}, z; \theta)- \log q(z|x_{i}; \phi)\bigg] = \sum_{i=1}^{N} \mathcal{L}(x_{i}; \phi, \theta),
\end{equation}
where $\mathds{E}_{q_{\phi}}\bigg[ \cdot \bigg] = \mathds{E}_{q(z|x_{i}; \phi)}\bigg[ \cdot \bigg]$ is the expectation with respect to the variational posterior $q(z|x_{i}; \phi)$.

Based on randomly drawn mini-batches of size $M$, we can re-construct the ELBO for amortized VI formulated in Equations \ref{equation:elbo_avi} as:
\begin{equation}
\label{equation:elbo_avi_1}
    \mathcal{L}(x; \phi, \theta) \simeq \frac{N}{M}\sum_{i=1}^{M}\mathcal{L}(x_{i}; \phi, \theta) = \frac{N}{M}\mathcal{\hat{L}}(x^{M}; \phi, \theta).
\end{equation}
Similar to the ELBO formulation in SVI (see Equation \ref{equation:svi}), the ELBO estimate based on $M$ sub-sample of the data is $\sum_{i=1}^{M}\mathcal{L}(x_{i}; \phi, \theta)$ which we denote by $\mathcal{\hat{L}}(x^{M}; \phi, \theta)$.
The optimization process for amortized VI (shown in Algorithm \ref{algorithm:avi}) usually follows the stochastic gradient ascent to ensure faster convergence.    

\vspace*{.2cm}
\begin{algorithm}[t]
 \caption{The amortized VI optimization process using stochastic gradient ascent}
 \label{algorithm:avi}
  \KwIn{Data \(x_{1:N}\)}
  \KwOut{Variational parameter \(\phi\); generative model parameter $\theta$}
  $\phi,  \theta \gets$ random initialization\\
  \While{$\mathcal{L}(x; \phi, \theta)$ has not converged}{
  \Repeat{all $N$ data points are seen at least once}{
    $\mathcal{M} \gets \{x^{M} | x^{M} \sim x_{1:N}$ and $|x^{M}| = M\}$ \\
    \For{$x^{M} \in \mathcal{M}$}{
    \For{$x_{j} \in x^{M}$}
    {
      Compute $\mathcal{L}(x_{j}; \phi, \theta)$\\
      Compute $\nabla_{\phi}\mathcal{L}(x_{j}; \phi, \theta); \nabla_{\theta}\mathcal{L}(x_{j}; \phi, \theta)$ \\
    }
    $\hat{\nabla}_{\phi}\mathcal{L}(x^{M}; \phi, \theta) = \frac{1}{M}\sum\limits_{j} \nabla_{\phi}\mathcal{L}(x_{j}; \phi, \theta)$ \\
    $\hat{\nabla}_{\theta}\mathcal{L}(x^{M}; \phi, \theta) = \frac{1}{M}\sum\limits_{j} \nabla_{\theta}\mathcal{L}(x_{j}; \phi, \theta)$ \\

    $\phi;\theta \gets$ Update parameters using $\hat{\nabla}_{\phi}\mathcal{\hat{L}}(x^{M}; \phi, \theta); \hat{\nabla}_{\theta}\mathcal{\hat{L}}(x^{M}; \phi, \theta)$\\
    $\displaystyle \mathcal{\hat{L}}(x^M; \phi, \theta) = \sum_{j=1}^{M}\mathcal{L}(x_{j}; \phi, \theta)$
    }
    }
    $\displaystyle \mathcal{L}(x; \phi, \theta) \simeq \frac{N}{M}\mathcal{\hat{L}}(x^{M}; \phi, \theta)$
}
\Return{$\phi$; $\theta$}
\end{algorithm}
\vspace*{.2cm}

However, using stochastic gradients does not guarantee an optimal solution as the gradient updates follow the steepest ascent in a Euclidean space without considering the parameter space's information geometry.
Natural gradients offer a solution to this problem as they reformulate the criterion for the gradient updates using the inverse of the Fisher Information matrix. 
With the use of deep learning models, comprising millions of parameters, in the form of the stochastic function, this computation for the inverse is infeasible as it has a time complexity of $\mathcal{O}(d^3)$ with $d$ being the dimension of the parameter space.
As discussed in Section \ref{section:natural_grad}, a simple trick would be to use the Hessian of the gradients to compute the Fisher Information matrix and subsequently, its inverse.
The Hessian can be computed using methods such as automatic differentiation or the reparameterization trick \shortcite{Khan2018FastAS}.
It is, however, not common to compute Hessians for deep models due to its high computational cost \shortcite{KhanN18}.

The Hessian computation can be avoided using the classical Gauss-Newton method \shortcite{schraudolph,martens_grad_hess}, in which the Hessian is approximated as the second moment of the gradients. 
The optimizer Adam \shortcite{Kingma2015AdamAM} can be used for this purpose as it computes the first and second moment of the gradients.
Further simplification in computation can be achieved by limiting the second moment to be a diagonal matrix, thereby enabling the computation for the inverse Fisher Information matrix to be $\mathcal{O}(d)$ rather than $\mathcal{O}(d^3)$, making it easy to apply to large deep learning problems.

\subsection{Variational Auto-Encoder (VAE)}
\label{subsection:vae}

\begin{figure}[ht]
    \centering
    \includegraphics[trim = 180 40 180 60, clip, width=0.4\columnwidth]{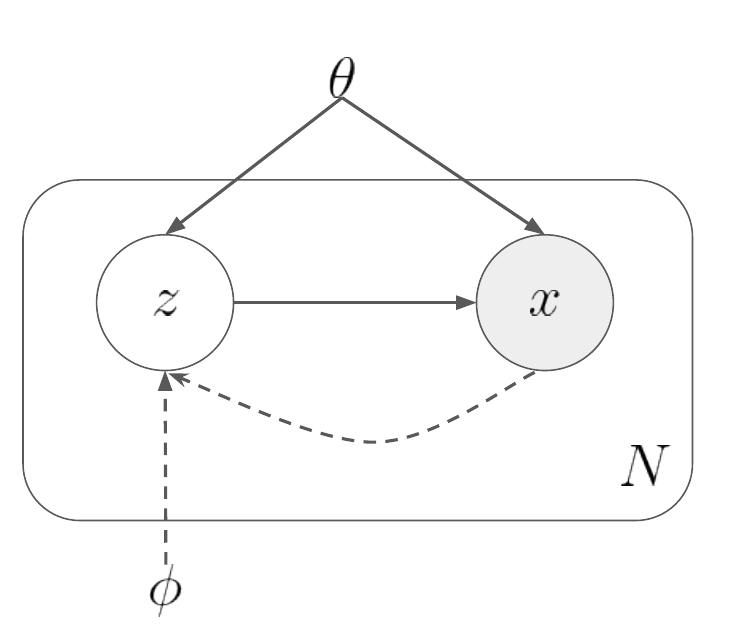}
    \caption{ Graphical model for the VAE framework. Solid lines denote the generative model, dashed lines denote the variational approximation to the intractable posterior density. The variational parameters $\phi$ are learned jointly with the generative model parameters $\theta$ \shortcite{kingma2013auto}.
    }
    \label{fig:vae}
\end{figure}

The VAE framework, developed by \shortciteA{kingma2013auto} and \shortciteA{RezendeMW14}, is an example of a statistical model that combines deep neural networks with the amortized VI optimization process.
VAEs employ two deep neural networks:
a probabilistic decoder, i.e., a top-down generative model that creates a mapping from a latent variable $z_{i}$ to a data point $x_{i}$, and a probabilistic encoder, i.e., a bottom-up inference model that approximates the posterior probability density, $p(z|x; \theta)$.
Correspondingly, these networks are commonly referred to as \textit{generative} and \textit{recognition} networks, respectively.
The graphical model for the VAE framework \shortcite{kingma2013auto} is illustrated in Figure \ref{fig:vae}.

We can get an intuitive understanding of the ELBO for VAEs by further re-arranging the terms of Equation \ref{equation:elbo_avi} as:
\begin{align}
\label{equation:elbo_vae}
\mathcal{L}(x; \phi, \theta) = \sum_{i=1}^{N} \mathcal{L}(x_{i}; \phi, \theta) &= \sum_{i=1}^{N} \mathds{E}_{q_{\phi}}\bigg[\log p(x_{i}, z; \theta) - \log q(z|x_{i}; \phi) \bigg], \nonumber \\
&= \sum_{i=1}^{N} \mathds{E}_{q_{\phi}}\bigg[\log p(x_{i}| z; \theta) + \log p(z;\theta) - \log q(z|x_{i}; \phi) \bigg], \nonumber \\
&= \sum_{i=1}^{N} \Bigg[\mathds{E}_{q_{\phi}}\bigg[\log p(x_{i}| z;\theta) \bigg] - D_{\text{KL}}\big(q(z|x_{i}; \phi)\;\|\;p(z; \theta)\big) \Bigg].
\end{align}
Thus, Equation \ref{equation:elbo_vae} establishes ELBO to be the sum of the expected log-likelihood and the negative KL-divergence between the approximate density and the prior over the latent variable evaluated at individual data points. 
The KL-divergence term can then be interpreted as regularizing $\phi$, encouraging the approximate posterior to be close to the prior $p(z; \theta)$ \shortcite{kingma2013auto}.
Furthermore, Equation \ref{equation:elbo_vae} establishes a connection to auto-encoders, as the first term is an expected negative reconstruction error while the KL-divergence term acts as a regularizer. 

\shortciteA{kingma2013auto} showed that applying the reparameterization trick to the ELBO formulation for VAEs in Equation \ref{equation:elbo_vae} yields the Stochastic Gradient Variational Bayes (SGVB) estimator $\mathcal{\Tilde{L}}^{B}(x; \phi, \theta) \simeq \mathcal{L}(x; \phi, \theta)$:
\begin{equation}
\label{equation:sgvb_vae}
    \mathcal{\Tilde{L}}^{B}(x; \phi, \theta) = \sum_{i=1}^{N} \Bigg[ \frac{1}{K} \sum_{k=1}^{K}\bigg[\log p(x_{i}|z_{(i,k)}; \theta) \bigg] - D_{\text{KL}}\big(q(z|x_{i}; \phi)\;\|\;p(z; \theta)\big) \Bigg].
\end{equation}
Often the KL-divergence term in Equation \ref{equation:sgvb_vae} can be integrated analytically such that only the expected reconstruction error requires estimation by sampling \shortcite{kingma2013auto}.
For the variational posterior, VAEs employ mean field approximation and as a simplifying choice, it is chosen to be a multivariate Gaussian with diagonal covariance structure:
\begin{equation*}
    q(z|x_{i}; \phi) = \prod_{j=1}^{J}q(z_{j}|x_{i}; \phi), \quad \log q(z|x_{i}; \phi) = \log \mathcal{N}(z; \mu_{i}, \sigma_{i}^2\mathds{I}),
\end{equation*}
where $J$ is the dimensionality of $z$, the mean, $\mu_{i}$, and standard deviation, $\sigma_{i}$, are the outputs of the encoder, i.e. non-linear functions of data point $x_{i}$ and the variational parameters $\phi$, which summarizes the corresponding neural network parameters \shortcite{kingma2013auto,zhang_vi_2019}.

As discussed in Section \ref{section:reparam}, the posterior is sampled as:
\begin{equation*}
    z_{(i,k)} \sim q(z|x_{i}; \phi) \quad\text{using}\quad z_{(i,k)} = g_{\phi}(x_{i}, \epsilon_{(i,k)}) = \mu_{i} + \sigma_{i} \odot \epsilon_{(i,k)},
\end{equation*}
where $\epsilon_{k} \sim \mathcal{N}(0, \mathrm{I})$ and $\odot$ denotes element-wise product.

Usually for the prior, a multivariate normal is chosen so that the latent variable $z$ can be drawn as:
\begin{equation*}
    z = \mathcal{N}(0, \mathds{I}).
\end{equation*}
However, a standard normal prior often leads to an over-regularization of the approximate posterior, which results in a less informative learned latent representation of the data \shortcite{nutan_flat}.
Recent advancements have shown to improve the representation learning in VAEs by modeling the prior to be dependent on additional parameters.
Finally,  the log-likelihood is computed as:
\begin{equation*}
    \log p(x_{i}|z_{(i,k)}; \theta) = \log p(x_{i}|\mu_{i}+\sigma_{i}\odot\epsilon_{(i,k)}; \theta),
\end{equation*}
signifying that the decoder network, parameterized by $\theta$, generates a data point $x_{i}$ through non-linear transformations of the latent vector $z_{i}$.
Thus, the resulting SGVB estimator for VAE is:
\begin{equation}
\label{equation:sgvb_vae_2}
    \mathcal{\Tilde{L}}^{B}(x; \phi, \theta) \simeq \sum_{i=i}^{N}\bigg[ \frac{1}{2}\sum_{j=1}^{J}\bigg( 1 + \log(\sigma_{(i,j)}^2)  - \mu_{(i,j)}^2 - \sigma_{(i,j)}^2 \bigg) + \frac{1}{K}\sum_{k=1}^{K}\log p(x_{i}|\mu_{i}+\sigma_{i}\odot\epsilon_{(i,k)}; \theta)\bigg],
\end{equation}
where $\mu_{(i,j)}$ and $\sigma_{(i,j)}$ denote the variational mean and standard deviation for the $j$-th element of these vectors evaluated at data point $x_{i}$, respectively. This formulation allows the stochastic estimate of the ELBO to be differentiated with respect to both $\phi$ and $\theta$ for gradient estimation.

In order to obtain a tighter ELBO and hence better variational approximations, importance sampling can be used to get a lower variance estimate of the evidence \shortcite{thin2021monte}.
This technique also forms the basis of the Importance Weighted Auto-Encoder (IWAE) \shortcite{iwae_burda} where the ELBO is computed as:
\begin{equation}
    \mathcal{L}(x; \phi, \theta)= \sum_{i=1}^{N}\mathds{E}_{q_{\phi}}\bigg[\log \frac{1}{K}\sum_{k=1}^{K} \bigg[\frac{p(x_{i}|z_{k}; \theta)p(z; \theta)}{q(z_{k}|x_{i}; \phi)} \bigg] \bigg],
\end{equation}
which is a $K$-sample importance weighting estimate of the log evidence.
\shortciteA{iwae_burda} showed that the true marginal likelihood is approached as $K \rightarrow \infty$.
However, \shortciteA{Nowozin2018DebiasingEA} proved that IWAEs introduce a biased estimator for the true marginal likelihood where the bias is reduced at a rate of $\mathcal{O}(1/K)$.
In addition, importance sampling is known to perform poorly in high dimensions \shortcite{mackay_book}.
To address these issues, \shortciteA{thin2021monte} proposed the Langevin Monte Carlo VAE (L-MCVAE), based on Sequential Importance Sampling (SIS), that provides a tighter ELBO than standard techniques as well as an unbiased estimator for the evidence.
However, \shortciteA{rainforth18b} provided empirical evidence that increasing the tightness of the ELBO independently to the expressive capacity of the recognition network can prove detrimental to its learning process.
In their study, \shortciteA{rainforth18b} demonstrated that deviating from this conventional training approach, 
specifically, by employing tighter bounds for training generative networks and looser bounds for recognition networks, more accurate posterior approximations and enhanced generative performance could be achieved.
Thus, they proposed three algorithms namely the Partially Importance Weighted Auto-Encoder (PIWAE), the Multiply Importance Weighted Auto-Encoder (MIWAE),
and the Combination Importance Weighted Auto-Encoder (CIWAE).
Each of these algorithms represents improvements over the IWAE and encompasses the standard IWAE as a special case.
\shortciteA{rainforth18b} evaluated the performance of these algorithms in terms of their ability to balance the training objectives between the recognition and the generative networks. 
While MIWAE and CIWAE primarily enabled achieving this balance, it was PIWAE that exhibited distinctive characteristics.
PIWAE not only successfully balanced the training objectives between their recognition and generative networks but also demonstrated simultaneous improvements in the training of both networks.
By focusing on improving the training of its recognition network, PIWAE indirectly influenced and heightened its generative network performance, while resulting in more accurate posterior approximations.

\subsection{Caveats and Solutions}

Although amortizing the inference contributes toward making the VI optimization faster and more scalable, it introduces certain issues. 
In this section, we describe pertinent issues such as the amortization gap, inconsistent representation learning in VAEs, the generalization gap, and the problem of posterior collapse.
Additionally, we cover the various methods that have been proposed in recent years to solve these issues.

\subsubsection{Sub-Optimal Inference}
\label{subsection:avi_gap}

In VI, the typical choice in the variational family is either factorized independent Gaussians or other mean field approximations for ease of analytical computation.
However, this limits the expressibility of the variational approximation by ignoring local dependencies between latent variables.
This limiting nature of the variational methodology results in the \textit{approximation gap} which can be reduced by choosing a family of variational densities that is flexible enough to contain the true posterior as one solution \shortcite{rezende2015variational}.

For this purpose, \shortciteA{rezende2015variational} proposed the concept of normalizing flow \shortcite{tabak_normal_flow,tabak_normal_flow_2} as a means to improve upon the expressiveness of the variational approximation.
A normalizing flow describes the transformation of a probability density through a sequence of invertible mappings \shortcite{rezende2015variational}.
It involves repeatedly applying change of variables to transform the simple initial variational probability density into a richer approximation to better match the true posterior density.
The main idea is to consider an invertible, smooth mapping $f : \mathds{R}^d \rightarrow \mathds{R}^d$ with inverse $f^{-1} = g$, such that $g(f(z)) = z$.
Thus, a random variable $z \sim q(z)$ can be transformed using the invertible, smooth function $f$ into a new random variable $z' = f(z)$ with density $q(z')$ as:
\begin{equation}
\label{equation:normaol_flow}
    q(z') = q(z)| \frac{\partial f^{-1}(z')}{\partial z'}| = q(z) | \frac{\partial f(z)}{\partial z} |^{-1},
\end{equation}
which is obtained using change of variables. 
With an appropriate choice of the transformation function, such that $| \frac{\partial f}{\partial z}|$ is easily computable, and applying Equation \ref{equation:normaol_flow} successively, complex and multi-modal densities can be efficiently constructed from simple factorized distributions such as independent Gaussians. 
In addition, different variants such as Langevin and Hamiltonian flows, invertible linear-time transformations as well as autoregressive flow \shortcite{chenkingma} have been proposed based on the concept of normalizing flows.

Alternatively, capturing the dependencies between latent variables increases the expressiveness of the variational family which mean field approximations, though effective in VI, discard.
The idea of auxiliary variables has been employed in hierarchical variational models (HVMs) \shortcite{ranganath_hvm_16} where dependencies between latent variables are induced similarly to the induction of dependencies between data in hierarchical Bayesian models.

Furthermore, in the case of amortized VI, using a stochastic function to estimate the variational density parameters instead of optimizing for each data point introduces a coding efficiency gap known as the \textit{amortization gap} \shortcite{Cremer2018InferenceSI}.
While offering significant benefits in computational efficiency, standard amortized inference models can suffer from sizable amortization gaps \shortcite{Krishnan2018OnTC}.
On the one hand, where the complexity of the variational density determines the approximation gap, it is the capacity of the stochastic function that results in the amortization gap.
The approximation gap, along with the amortization gap, contributes toward the \textit{inference gap}, which is the gap between the marginal log-likelihood and the ELBO.

In their work, \shortciteA{Cremer2018InferenceSI} observed that for VAEs, trained especially on complex data sets, the amortization gap contributes significantly towards the inference gap.
They combined normalizing flow with the induction of hierarchical auxiliary variables to increase the expressiveness of the variational approximation.
This resulted in generalizing inference in addition to improving the complexity of the variational approximation.
\shortciteA{Cremer2018InferenceSI} demonstrated through their experiments that increasing the capacity of the encoder reduces the amortization gap.
However, \shortciteA{Shu2018AmortizedIR} argue that an over-expressive encoder degrades generalization.
Therefore, in their paper, \shortciteA{Shu2018AmortizedIR} introduced the concept of amortized inference regularization which is a regularization technique that restricts the capacity of the encoder to prevent both the inference and the generative models from over-fitting to the training set (explained in detail in Section \ref{subsect:gen_gap}).

A recent research trend has seen an effort toward reducing the amortization gap using an iterative training approach.
For instance, \shortciteA{hjelm_iterative} proposed a training procedure to iteratively refine the chosen approximate posterior estimated by a recognition network.
The proposed learning algorithm follows expectation-maximization (EM), wherein the E-step the recognition network is used to initialize the parameters of the variational posterior which are then iteratively refined. 
This refinement procedure provides a tight and asymptotically unbiased estimate of the ELBO,
which is used to train both the recognition and generative models during the M-step. 
Moreover, this refinement procedure results in lower variance Monte Carlo estimates for the approximate posterior and provides a more accurate estimate of the log-likelihood of the model \shortcite{hjelm_iterative}.
On a similar note, \shortciteA{Marino2018IterativeAI} proposed an iterative training scheme that reduces the amortization gap in standard VAEs by directly encoding the gradients of the parameters of the approximated posterior.
VAEs create direct, static mappings from observations to the parameters of the approximate posterior with the optimization of these parameters replaced with the optimization of a shared, i.e., amortized, set of parameters $\phi$ for the recognition model \shortcite{Marino2018IterativeAI}.
This optimization process makes the recognition network in a VAE a purely bottom-up inference process which does not correspond well to perception \shortcite{ladder_vae}.
In other words, inference is as much a top-down as it is a bottom-up process, and therefore, in order to combine the two, \shortciteA{Marino2018IterativeAI} proposed a training regimen that enables a VAE to learn to perform inference by iteratively encoding the gradients of the approximate posterior parameters, which are rarely performed in practice.
The results from their experiments showed that this form of iterative training continuously refined the approximate posterior estimate, thereby, reducing the amortization gap.
However, this method also required additional computation over VAEs with similar architectures.

A semi-amortized approach proposed by \shortciteA{Kim2018SemiAmortizedVA} is another iterative training approach that used amortized variational inference to initialize the variational parameters and then subsequently ran SVI procedure for local iterative refinement of these parameters.
The resulting Semi-Amortized VAE (SA-VAE) framework had a smaller amortization gap than vanilla VAEs. 
Additionally, it avoided the posterior collapse phenomenon, common in VAEs, wherein the variational posterior collapses to the prior (discussed in detail in Section \ref{subsection:posterior_collapse}).
However, this semi-amortized approach suffered from additional computation overhead, owing to the additional SVI optimization at test time. 
In order to tackle this issue, \shortciteA{Kim2021ReducingTA} proposed an approach that aimed at reducing the amortization gap by considering this difference between the true posterior and amortized posterior distribution as random noise. 
They showed that this approach is more efficient than the recent semi-amortized approaches, being able to perform a single feed-forward pass during inference.

\subsubsection{Inconsistent Representation Learning}
\label{subsection:repr}

VAEs are powerful frameworks that make use of amortized VI for unsupervised learning.
Amortizing the inference enables VAEs to perform scalable variational posterior approximation in deep latent variable models.
As discussed in Section \ref{subsection:vae}, VAEs amortize the posterior inference by the use of a stochastic function that maps observations to their subsequent representations in the latent space.
Once trained, the recognition model of a VAE can be used to obtain low-dimensional representations of data, the quality of which determines the applicability of VAEs.
However, the recognition model of a fitted VAE tends to map an observation and its subsequent semantics-preserving transformation (e.g., rotation, translation) to different parts in the latent space \shortcite{sinha2021consistency}.
This inconsistency of the recognition network has an adverse effect on the quality of the learned representations as well as generalization.
To enforce consistency in VAEs, \shortciteA{sinha2021consistency} proposed a regularization technique; the idea of which is to minimize the KL-divergence between the variational approximations when conditioned on an observation and when conditioned on its randomly transformed semantics-preserving counterpart.
\shortciteA{sinha2021consistency} termed the resulting VAE trained with this regularization technique as the consistency-regularized VAE (CR-VAE).

Based on the formulation of the ELBO for amortized VI from Equation \ref{equation:elbo_avi}, \shortciteA{sinha2021consistency} defined a semantics-preserving transformation distribution $t(\Tilde{x}|x_{i})$ for a data point $x_{i}$ with the argument that a vanilla VAE, once fit to data, fails to output similar latent representations for both $x_{i}$ and $\Tilde{x}_{i}$ in comparison to a good representation learning algorithm.
The CR-VAE addresses this issue by re-defining the ELBO objective for VAEs as:
\begin{equation}
\label{equation:elbo_crvae}
    \mathcal{L}_{\text{CR-VAE}}(x; \phi, \theta) = \mathcal{L}(x; \phi, \theta) + \sum_{i=1}^{N}\bigg[\mathds{E}_{t(\Tilde{x} | x_{i})}\bigg[\mathcal{L}(x_{i}; \phi, \theta)\bigg] - \lambda \cdot \mathcal{R}(x_{i}; \phi)\bigg],
\end{equation}
s.t.,
\begin{equation*}
    \Tilde{x}_{i} \sim t(\Tilde{x}|x_{i}) \Longleftrightarrow \epsilon \sim \mathcal{U}[-\delta, \delta] \quad\text{and}\quad \Tilde{x}_{i} = g(x_{i}, \epsilon).
\end{equation*}
The function $ g(x_{i}, \epsilon)$ is a semantics-preserving transformation of a data point $x_{i}$, e.g., translation with a random length $\epsilon$ drawn from a uniform distribution $\mathcal{U}[-\delta, \delta]$ for some threshold $\delta$ \shortcite{sinha2021consistency}.
Additionally, the final term in Equation \ref{equation:elbo_crvae} is the regularization term which is defined as:
\begin{equation*}
    \mathcal{R}(x_{i}; \phi) = \mathds{E}_{t(\Tilde{x} | x_{i})}\bigg[D_{\text{KL}}(q(z|\Tilde{x}_{i}; \phi)\;\|\; q(z|x_{i}; \phi))\bigg].
\end{equation*}
Maximizing the objective in Equation \ref{equation:elbo_crvae} maximizes the likelihood of the data and their augmentations while enforcing consistency through $\mathcal{R}(x_{i}; \phi)$ \shortcite{sinha2021consistency}. 
The regularization term only affects the recognition model (with parameters $\phi$), and minimizing it forces the representations of each observation and their corresponding augmentations to lie close to each other in the latent space. 
The strength of the regularizer is controlled by the hyper-parameter $\lambda \geq 0$.

Through their experiments on the MNIST \shortcite{mnist}, Omniglot \shortcite{omniglot}, and CelebA \shortcite{celebA} data sets, \shortciteA{sinha2021consistency} showed that CR-VAE improved the learned representations over vanilla VAEs and improved generalization performance.
Additionally, they applied the regularization technique to IWAE \shortcite{iwae_burda}, $\beta$-VAE \shortcite{higgins2016beta}, and nouveau VAE \shortcite{nvae} and demonstrated that CR-VAEs yielded better representations and generalized performance than their base VAEs.

Further research in this direction were conducted to show that vanilla VAE models are not \textit{auto-encoding} \shortcite{cemgil2020autoencoding}, i.e., samples from the generative network are not mapped to corresponding representations by the recognition network.
\shortciteA{cemgil2020autoencoding} derived the Auto-encoding VAE (AVAE) framework that utilizes a reformed lower bound to achieve adversarial robustness for the learned representations.
In addition, an AVAE model optimized with this lower bound facilitates data augmentations and self-supervised density estimation.
The central idea of AVAE is making the recognition and the generative networks to be consistent both on the training data and on the auxiliary observations generated by the generative network \shortcite{cemgil2020autoencoding}.
Through their experiments on the colourMNIST and CelebA data sets, \shortciteA{cemgil2020autoencoding} showed that their proposed AVAE framework, using both multi-layered perceptron and convolutional neural network architectures, achieved high adversarial accuracy without adversarial training.

\subsubsection{Generalization Gap}
\label{subsect:gen_gap}

On examining the amortized VI formulation for the ELBO from Equation \ref{equation:elbo_vae}, \shortciteA{Shu2018AmortizedIR} concluded that it is a \textit{data-dependent regularized} maximum likelihood objective, which is a means to restrict the recognition model capacity.
While a low-capacity recognition model increases the amortization gap, an over-expressive one harms generalization.
This amortized inference regularization (AIR) strategy encourages recognition model smoothness while reducing the inference gap and increasing the log-likelihood on the test set. 
\shortciteA{Shu2018AmortizedIR} proposed a modification to the vanilla VAE by injecting noise into the recognition model resulting in the denoising VAE (DVAE).
Although DVAEs were originally proposed by \shortciteA{denoise_bengio}, \shortciteA{Shu2018AmortizedIR} further demonstrated that the optimal DVAE model is a kernel regression model, and the variance of the injected noise controls the smoothness of the optimal recognition model.
Additionally, \shortciteA{Shu2018AmortizedIR} proposed the weight-normalized inference (WNI) method which leverages the weight normalization technique introduced by \shortciteA{Salimans_weight}, to control the capacity and the smoothness of the recognition model.
Through their experiments on the Caltech 101 Silhouettes \shortcite{caltech101_silhouettes} and statically binarized MNIST and Omniglot data sets, \shortciteA{Shu2018AmortizedIR} showed that regularizing the recognition either by the DVAE or the WNI-VAE method improved the test set log-likelihood performance. 
From the results on these data sets, a consistent reduction of the test set inference gap was noticed when the inference model was regularized. 

As discussed in Section \ref{subsection:vae}, the VAE amortizes the inference to scale its training to large data sets, making it a popular choice for several applications such as density estimation, lossless compression, and representation learning \shortcite{zhanggeneralization}.
However, the use of amortized inference during its training phase can lead to poor generalization performance.
In order to tackle this issue, \shortciteA{zhanggeneralization} introduced a training methodology for the recognition network in a VAE to reduce over-fitting to the training data and hence, improve generalization.
Due to the lack of sufficient training data, a flexible posterior approximation can lead the recognition network to reduce the overall inference gap but also over-fit to the training data.
\shortciteA{zhanggeneralization} proposed a self-consistent training method wherein a mixture of samples from the training data set and those generated by the generative model were fed to the recognition network during the training phase.
This mixture of distributions could be interpreted as a form of training data augmentation to help overcome the over-fitting caused by the application of amortized inference \shortcite{zhanggeneralization}.
The results from their experiments showed that this training approach consistently improved the generalization performance, as measured by the negative ELBO on both the binary and grey-scale MNIST data sets \shortcite{binary_mnist,mnist}.

\subsubsection{Posterior Collapse}
\label{subsection:posterior_collapse}

Posterior collapse is a phenomenon often observed in VAE training, which arises when the variational posterior distribution lies close, or as the name suggests, collapses, to the prior. 
This causes the generative network to ignore a subset of the latent variables.  
The model, hence, fails to learn a valuable representation of the data.

Several works \shortcite{Yang2017Improved,semeniuta2017Hybrid,zhao_song_ermon_2019,tolstikhin2018wasserstein,takida2021arelbo} suggest this phenomenon stems from two main reasons. 
First, in cases where the generative network is especially powerful, it models the observed variable $x$ independently, causing the latent variables $z$ to get ignored. 
Second, with the training objective of maximizing the ELBO and minimizing the KL-divergence term, as observed in Equation \ref{equation:elbo_vae}, the variational posterior collapses to the prior as the KL-divergence term approaches zero. 
Moreover, \shortciteA{lucas2019} suggested that this occurs due to the spurious local maxima in the training objective instead.

Various approaches tackle the posterior collapse by either replacing the generative network with a weaker capacity alternative \shortcite{Yang2017Improved,semeniuta2017Hybrid}, or by modifying the ELBO training objective \shortcite{zhao_song_ermon_2019,tolstikhin2018wasserstein,takida2021arelbo}.
\shortciteA{takida2021arelbo} demonstrated that inconsistency in choosing certain hyperparameters, more specifically data variance parameters, leads to over-smoothing and, in turn, posterior collapse. They proposed an adaptively regularized ELBO objective function to control the model smoothing and posterior collapse.
\shortciteA{semeniuta2017Hybrid}, on the other hand, proposed replacing the traditional Recurrent Neural Networks for text generation with a weaker convolution-deconvolution architecture, which results in faster convergence as well as forces the network to learn from the latent dimensions.

Although these approaches successfully tackle posterior collapse, they either require an alteration to the training objective or do not fully utilize the recent advances in deep auto-regressive networks.  
Alternatively, \shortciteA{razavi2019} proposed $\delta$-VAEs, which still leverages deep auto-regressive networks and the training objective while enforcing a minimum KL-divergence between the variational posterior and prior. 
\shortciteA{ZhuBLMLW20} demonstrated that considering the KL-divergence for the entire data set distribution instead of a single data point is enough to reduce posterior collapse by keeping a positive expectation. 
They additionally proposed a Batch-Normalized VAE to set a lower bound on the expectation.

\section{Beyond KL-divergence}
\label{section:beyond}

The KL-divergence offers a computationally convenient solution to measure the dissimilarity between two distributions.
As discussed in Section \ref{susbsection:kld}, a closed-form solution for the forward KL-divergence is unavailable in the case of VI, and therefore, the reverse KL-divergence is used to formulate the VI objective function.
The reverse KL-divergence, also known as I-projection or information projection \shortcite{Murphy1991}, in the case of amortized VI, is formulated as:
\begin{equation}
    D_{\text{KL}}(q(z|x_{i}; \phi)\;\|\; p(z|x_{i}; \theta)) = \mathds{E}_{q}\bigg[\log \frac{q(z|x_{i}; \phi)}{p(z|x_{i}; \theta)} \bigg],
\end{equation}
s.t.,
\begin{equation*}
    \lim_{p(z|x_{i}; \theta)\to 0} \frac{ q(z|x_{i}; \phi)}{p(z|x_{i}; \theta)} \to \infty \; \quad\text{where}\quad \;  q(z|x_{i}; \phi) > 0.
\end{equation*}
The limit indicates the need to force $q(z|x_{i}; \phi) = 0$ wherever $p(z|x_{i}; \theta) = 0$, otherwise the KL-divergence would be very large \shortcite{ganguly2021introduction}. 
This \textit{zero forcing} nature of the reverse KL-divergence has been proven to be useful in settings such as multi-modal posterior densities with unimodal approximations \shortcite{chi_vi}.
In such cases, the zero forcing nature helps to concentrate on one mode rather than spread mass all over them \shortcite{10.5555/1162264}.
However, zero forcing leads to an underestimate of the posterior variance \shortcite{chi_vi}.
In addition, it leads to degenerate solutions during optimization and is the source of \textit{pruning} in VAEs \shortcite{chi_vi,iwae_burda}.
As a result of these shortcomings, several other divergence measures have been proposed in recent years.
In this section, we discuss a few of the relevant divergence measures and how they are used in the context of VI.

\subsection{$\chi$-divergence}

\shortciteA{chi_vi} proposed CHIVI, a VI algorithm that minimizes the $\chi$-divergence between the variational approximation and the true posterior density.
For amortized VI, the divergence measure is defined as:
\begin{equation}
\label{equation:chi_div}
    \mathcal{D}_{\chi^r} = D_{\chi^r}(p(z|x_{i}; \theta) \;\|\; q(z|x_{i}; \phi)) = \mathds{E}_{q_{\phi}}\bigg[\bigg(\frac{p(z|x_{i}; \theta)}{q(z|x_{i}; \phi)}\bigg)^r -1 \bigg],
\end{equation}
where $r$ is chosen depending on the application and data set.

Optimizing Equation \ref{equation:chi_div} leads to a variational density with \textit{zero avoiding} behavior like the forward KL-divergence \shortcite{Murphy1991} or expectation propagation (EP) \shortcite{Minka2005DivergenceMA}.
This indicates that the $\chi$-divergence is infinite whenever $q(z|x_{i}; \phi) = 0$ and $p(z|x_{i}; \theta) > 0$ and thus, minimizing the $\chi$-divergence while setting $p(z|x_{i}; \theta) > 0$ forces $q(z|x_{i}; \phi) > 0$ \shortcite{chi_vi}.
Therefore, $q$ avoids allocating zero mass at locations where $p$ has non-zero mass.
In contrast to VI optimization that uses KL-divergence as a means to maximize a lower bound on the model evidence, the main idea behind CHIVI is to optimize an upper bound which \shortciteA{chi_vi} refer to as the $\chi$ \textit{upper bound} (CUBO).
Minimizing the CUBO is equivalent to minimizing the $\chi$-divergence \shortcite{chi_vi}.
The $\chi$-divergence objective function, for amortized VI, over $N$ data points can be formulated as:
\begin{align}
\label{equation:CUBO}
    \prod_{i=1}^{N}\bigg(\mathcal{D}_{\chi^r} + 1 \bigg) &= \prod_{i=1}^{N}\mathds{E}_{q_{\phi}}\bigg[\bigg(\frac{p(z|x_{i}; \theta)}{q(z|x_{i}; \phi)}\bigg)^r\bigg], \nonumber \\
    \sum_{i=1}^{N} \log \bigg(\mathcal{D}_{\chi^r} + 1 \bigg) &= \sum_{i=1}^{N} \log \mathds{E}_{q_{\phi}}\bigg[\bigg(\frac{p(z|x_{i}; \theta)}{q(z|x_{i}; \phi)}\bigg)^r\bigg], \nonumber \\
    &=\sum_{i=1}^{N} \log \mathds{E}_{q_{\phi}}\bigg[\bigg(\frac{p(z, x_{i}; \theta)}{p(x_{i}; \theta)q(z|x_{i}; \phi)}\bigg)^r\bigg], \nonumber \\
    &=-r\sum_{i=1}^{N} \log p(x_{i}; \theta) + \sum_{i=1}^{N} \log \mathds{E}_{q_{\phi}}\bigg[\bigg(\frac{p(z, x_{i}; \theta)}{q(z|x_{i}; \phi)}\bigg)^r\bigg], \nonumber \\
    &=-r\log p(x; \theta) + \sum_{i=1}^{N} \log \mathds{E}_{q_{\phi}}\bigg[\bigg(\frac{p(z, x_{i}; \theta)}{q(z|x_{i}; \phi)}\bigg)^r\bigg], \nonumber \\
    \log p(x; \theta) &= \frac{1}{r}\sum_{i=1}^{N} \log \mathds{E}_{q_{\phi}}\bigg[\bigg(\frac{p(z, x_{i}; \theta)}{q(z|x_{i}; \phi)}\bigg)^r\bigg] - \frac{1}{r}\sum_{i=1}^{N} \log \bigg(\mathcal{D}_{\chi^r} + 1 \bigg), \nonumber \\
    &= \text{CUBO}_{r} - \frac{1}{r}\sum_{i=1}^{N} \log \bigg(\mathcal{D}_{\chi^r} + 1 \bigg),
\end{align}
where
\begin{equation*}
    \text{CUBO}_{r} = \frac{1}{r}\sum_{i=1}^{N} \log \mathds{E}_{q_{\phi}}\bigg[\bigg(\frac{p(z, x_{i}; \theta)}{q(z|x_{i}; \phi)}\bigg)^r\bigg]
\end{equation*}
is a non-decreasing function of the order of the $\chi$-divergence $\forall r \geq 1$.
By non-negativity of the $\chi$-divergence in Equation \ref{equation:CUBO}, an upper bound to the log-likelihood of data is established as:
\begin{equation*}
    \log p(x; \theta) \leq \text{CUBO}_{r}.
\end{equation*}
When $r \geq 1$, $\text{CUBO}_r$ is an upper bound to the model evidence enabling a higher precision approximation of $\log p(x; \theta)$ as $r$ approaches $1$.  
\shortciteA{chi_vi} stated that the gap induced by $\text{CUBO}_{r}$ and ELBO increases with $r$; however, as $r$ decreases to $0$, $\text{CUBO}_{r}$ becomes a lower bound as tends to the ELBO, i.e., $\lim_{r \to 0} \text{CUBO}_r = \text{ELBO}$.

\subsection{$\alpha$-divergence}

The KL-divergence is a special case of a family of divergence measures known as the $\alpha$-divergence.
Different formulations of the $\alpha$-divergence exist \shortcite{alpha_div,inf_gen_zhu}; however, we focus on R{\'e}nyi's formulation, which defines the divergence measure for amortized VI as:
\begin{equation}
    \mathcal{D}_{\alpha} = D_{\alpha}(q(z|x_{i}; \phi)\;\|\; p(z|x_{i}; \theta)) = \frac{1}{\alpha-1} \log \int q(z|x_{i}; \phi)^{\alpha} p(z|x_{i}; \theta)^{1-\alpha} \mathrm{d}z,
\end{equation}
where $\alpha \in [0,1) \cup (1, \infty)$ and as $\alpha \to 1$, we recover the standard reverse KL-divergence for VI \shortcite{renyi_div}.
A special case of $\alpha=2$ results in a measure that is proportional to the $\chi^2$-divergence.

Using $\alpha$-divergence, a bound on the marginal likelihood can be derived as:
\begin{align}
\label{equation:alpha_bound}
    \mathcal{L}_{\alpha}(x; \phi, \theta) &= \log p(x; \theta) - \frac{1}{\alpha-1}\sum_{i=1}^{N} \mathcal{D}_{\alpha}, \nonumber \\
    &=\sum_{i=1}^{N}\log p(x_{i}; \theta) - \frac{1}{\alpha-1}\sum_{i=1}^{N} \mathcal{D}_{\alpha}, \nonumber \\
    &=\sum_{i=1}^{N}\bigg[\frac{\log p(x_{i}; \theta)^{\alpha-1}}{\alpha-1} - \frac{\mathcal{D}_{\alpha}}{\alpha-1} \bigg], \nonumber \\
    &=\frac{1}{\alpha-1}\sum_{i=1}^{N}\bigg[-\log p(x_{i}; \theta)^{1-\alpha} - \mathcal{D}_{\alpha}\bigg], \nonumber \\
    &=\frac{1}{1-\alpha}\sum_{i=1}^{N}\bigg[\log p(x_{i}; \theta)^{1-\alpha} + \log \int q(z|x_{i}; \phi)^{\alpha} p(z|x_{i}; \theta)^{1-\alpha} \mathrm{d}z \bigg], \nonumber \\
    &=\frac{1}{1-\alpha}\sum_{i=1}^{N}\bigg[\log p(x_{i}; \theta)^{1-\alpha} \int q(z|x_{i}; \phi)^{\alpha} p(z|x_{i}; \theta)^{1-\alpha} \mathrm{d}z \bigg], \nonumber \\
    &=\frac{1}{1-\alpha}\sum_{i=1}^{N}\bigg[\log \int q(z|x_{i}; \phi) p(x_{i}; \theta)^{1-\alpha} q(z|x_{i}; \phi)^{\alpha-1} p(z|x_{i}; \theta)^{1-\alpha} \mathrm{d}z \bigg], \nonumber \\
    &=\frac{1}{1-\alpha}\sum_{i=1}^{N}\log \mathds{E}_{q_{\phi}}\bigg[ p(x_{i}; \theta)^{1-\alpha}q(z|x_{i}; \phi)^{\alpha-1} p(z|x_{i}; \theta)^{1-\alpha} \bigg], \nonumber \\
    &=\frac{1}{1-\alpha}\sum_{i=1}^{N}\log \mathds{E}_{q_{\phi}}\bigg[q(z|x_{i}; \phi)^{\alpha-1} p(z, x_{i}; \theta)^{1-\alpha} \bigg], \nonumber \\
    &=\frac{1}{1-\alpha}\sum_{i=1}^{N}\log \mathds{E}_{q_{\phi}}\bigg[ \bigg(\frac{p(z, x_{i}; \theta)}{q(z|x_{i}; \phi)} \bigg)^{1-\alpha}\bigg].
\end{align}
The bound in Equation \ref{equation:alpha_bound}, also known as Variational R{\'e}nyi (VR) bound \shortcite{renyi_vi}, can be extended to $\alpha < 0$ and is continuous and non-increasing on $\alpha \in \{\alpha: |\mathcal{L}_{\alpha}(x)|< +\infty\}$ \shortcite{renyi_vi}.
Especially for all $0 < \alpha_{+} < 1$ and $\alpha_{-} < 0$,
\begin{equation*}
    \mathcal{L}(x; \phi, \theta) < \mathcal{L}_{\alpha_{+}}(x; \phi, \theta) < \log p(x; \theta) < \mathcal{L}_{\alpha_{-}}(x; \phi, \theta),
\end{equation*}
indicating that the VR bound can be useful for model selection by sandwiching the marginal
likelihood with bounds computed using positive and negative $\alpha$ values.
In their work, \shortciteA{renyi_vi} demonstrated how choosing different $alpha$ values allows the variational approximation to balance between zero forcing ($\alpha \to \infty$) and mass-covering ($\alpha \to -\infty$) behavior.

$\alpha$-divergences are a subset of a more general family of divergences known as $f$-divergences \shortcite{f_div}, which for amortized VI can be formulated as:
\begin{equation}
    D_{f}(q(z|x_{i}; \phi)\;\|\; p(z|x_{i}; \theta)) = \int p(z|x_{i}; \theta)f\bigg( \frac{q(z|x_{i}; \phi)}{p(z|x_{i}; \theta)}\bigg) \mathrm{d}z,
\end{equation}
where $f$ is a convex function with $f(1)=0$, and the reverse KL-divergence for the above formulation can be obtained by choosing the $f$-divergence as $f(\omega)=\omega\log(\omega)$.
In general, only specific choices of $f$ result in a bound that is trivially dependent on the marginal likelihood, and which is, therefore, useful for VI \shortcite{zhang_vi_2019}.

\subsection{Stein Discrepancy}

Introduced by \shortciteA{stein1972bound} as an error bound to measure how well an approximate distribution fits a distribution of interest, the Stein discrepancy as a divergence measure for amortized VI can be defined as:
\begin{equation}
\label{equation:stein}
    D_{\text{stein}}(p(z|x_{i}; \theta)\;\|\; q(z|x_{i}; \phi)) = sup_{f\in \mathcal{F}}\bigg[ \mathds{E}_{q_{\phi}}\bigg[f(z)\bigg] - \mathds{E}_{p_{\theta}}\bigg[f(z)\bigg]\bigg]^2,
\end{equation}
where $\mathcal{F}$ denotes a set of smooth, real-valued functions \shortcite{zhang_vi_2019}.
The smaller this divergence is, the more similar $p$ and $q$ are. 
When this divergence is zero, the two densities are identical.

The second term in Equation \ref{equation:stein} involves taking expectations with respect to the intractable posterior.
Therefore, in VI, the Stein discrepancy can only be used for classes of functions $\mathcal{F}$ for which the second term is zero.
A suitable class with this property can be defined by applying a differential operator $\mathcal{A}$ on an arbitrary smooth function $g$ as:
\begin{equation*}
    f(z) = \mathcal{A}g(z),
\end{equation*}
where $z \sim p(z)$ and the operator $\mathcal{A}$ are constructed in a way such that the second expectation in Equation \ref{equation:stein} is zero.
A popular choice for the operator that fulfills this requirement is the Stein operator, which is defined as:
\begin{equation*}
    \mathcal{A}g(z) = g(z)\nabla_{z}\log p(x_{i}, z; \theta) + \nabla_{z}g(z),
\end{equation*}
where $\nabla_{z}\log p(x_{i}, z; \theta)$ is the score function, which can be calculated without knowing the normalization constant of $p(x_{i},z; \theta)$.
This indicates the Stein discrepancy's robustness in handling model misspecifications in the form of unnormalized distributions \shortcite{liu_wang_stein}.

Several research in VI have used the Stein discrepancy in recent years \shortcite{stein_vi,liub16,liu_wang_stein,Liu2017SteinVP}.
Of particular interest is the Stein Variational Gradient Descent (SVGD) methodology, proposed by \shortciteA{liu_wang_stein}, which makes use of the kernelized Stein discrepancy (KSD) \shortcite{liub16} along with the Stein operator to construct the variational objective.
With a specific choice of the kernel and $q$, KSD enables the SVGD algorithm to determine the optimal perturbation in the direction of the steepest gradient of the KL-divergence \shortcite{liu_wang_stein}.
Similar to the normalizing flow approach \shortcite{tabak_normal_flow,tabak_normal_flow_2,rezende2015variational}, SVGD leads to a scheme where samples in the latent space are sequentially transformed to approximate the true posterior distribution.
Furthermore, the SVGD algorithm  does not require explicitly specified
parametric forms, allowing it to be flexible and easily implementable \shortcite{liu_wang_stein}.

Moreover, \shortciteA{liub16} demonstrated how the KSD acts as an unbiased statistic for measuring the goodness-of-fit test, which determines how likely it is that a set of given samples were generated from a target density function \shortcite{chwialkowski_stein}.
Thus, in conjunction with being formulated as a gradient operator in SGVD, the Stein discrepancy can equivalently be expressed as a test function formulation for goodness-of-fit tests.
This duality property of the Stein discrepancy is what makes it particularly useful in statistical inference and hypothesis testing.
However, both KSD and SVGD in their original form can only be applied to continuous distributions. 
Hence, in their work, \shortciteA{pmlr-v108-han20c} developed the gradient-free versions of both KSD and SVGD for goodness-of-fit testing on discrete distributions and to sample form discrete-valued distributions while performing approximate inference respectively.

\section{Open Problems}
\label{section:open}

In this paper, we provide a mathematical foundation for amortized VI through studying and gaining an intuitive understanding of traditional-, stochastic-, and black box-VI, and the strengths and weaknesses of these methods.
Additionally, we elucidate the recent advancements in the field of amortized VI, particularly in addressing its issues - sub-optimal inference, inconsistent representation learning, generalization gap, and posterior collapse.
Furthermore, we provide an overview of the various alternate divergence measures that can potentially replace the KL-divergence measure in the VI optimization process.
Although the use of amortized VI in deep generative modeling has grown in recent years, the research to make it scalable, efficient, accurate, and easier to formulate is still ongoing.
We outline some of the active research areas and open problems in the field of VI :

\begin{itemize}
\item \textbf{Amortized VI and Deep Learning}. With the recent advancements in the field of deep learning, researchers have successfully combined VI along with deep neural networks, in the form of VAEs, for generative modeling tasks.
However, VAEs lack the ability to take into account the uncertainty in posterior approximation in a principled manner \shortcite{Kim2021ReducingTA}. 
Recent research \shortcite<e.g.,>{NEURIPS2021_d5b03d3a,Shridhar2018BayesianCN} has been aimed towards making the posterior approximation in VAEs more interpretable by using Bayesian neural networks \shortcite{bayes_nn} as the choice for the parametric functions for both the inference and generative models in VAEs.

\item \textbf{VAE latent space geometry}. Generally, the distance between points in the latent space in a VAE does not reflect the true similarity of corresponding points in the observation space \shortcite{chen2018metrics}.
Notable research in this area includes treating the latent space of VAEs as a Riemannian manifold \shortcite{chen2018metrics}. 
Iterating on this idea, \shortciteA{nutan_flat} developed the \textit{flat manifold} VAE which defines the latent space as a Riemannian manifold and regularizes
the Riemannian metric tensor to be a scaled identity matrix.
This extension to vanilla VAEs allowed for learning flat latent manifolds, where the Euclidean distance is a proxy for the similarity between data points.
Although there has been some progress to improve the representation learning in VAEs (see Section \ref{subsection:repr}), the geometrical properties of the latent space in VAEs are not well understood.

\item \textbf{Alternatives to the non-convex ELBO}. Another area of research is to address the non-convex nature of the ELBO.
With the recent introduction of thermodynamic integration techniques \shortcite{lartillot_philippe}, researchers have paved the way for the development of a new VI framework that uses the Variational H\"{o}lder (VH) bound \shortcite{bouchard2015approximate} as an alternative to the ELBO.
Unlike the ELBO, the VH bound is a convex upper bound to the intractable evidence \shortcite{bouchard2015approximate}, the minimization of which is a convex optimization problem that can be solved using existing convex optimization algorithms.
Additionally, the approximation error of the VH bound can also be analyzed using tools from convex optimization literature \shortcite{bouchard2015approximate}.
Furthermore, promising work in this area by \shortciteA{chen2021variational} has shown to achieve a one-step approximation of the exact marginal log-likelihood using the VH bound.

\item \textbf{Automatic VI}. Finally, the development of probabilistic programming tools, such as PyMC \shortcite{Salvatier2016}, Stan \shortcite{Carpenter_stan}, Infer.Net \shortcite{InferNET18}, Zhusuan \shortcite{shi2017zhusuan}, have enabled researchers to automatize their experiment pipelines and thus allowing them to revise and improve their models with ease.
Despite the progress in the development of these toolboxes, their usage is not straightforward to researchers new to the field.
\end{itemize}

\acks{We are grateful to our colleagues Aubin Samacoits and Dhruba Pujary at Sertis Vision Lab for their constructive input and feedback during the writing of this paper.
}

\appendix
\section{Derivation of Equation \ref{equation:mean_field_1}}
\label{appendix:appendix_A}

\begin{align*}
    \mathcal{L}(x; \xi_{1:N}, \theta) &= \sum_{i=1}^{N} \mathds{E}_{q}\bigg[\log p(x_{i}, z; \theta) - \log q(z|x_{i}; \xi_{i}) \bigg], \\
    &= \sum_{i=1}^{N} \int \prod_{k}q_{k}\bigg[\log p(x_{i}, z; \theta) - \log \prod_{k}q_{k}\bigg]\mathrm{d}z.
\end{align*}
Now, we shall focus on the second term of the above equation as:
\begin{subequations}
\begin{align*}
    &\int \prod_{k}q_{k}\log \prod_{k}q_{k}\mathrm{d}z \\
    &=\int q_{j} \prod_{k \neq j } q_{k} \log \Big[ q_{j} \prod_{k \neq j } q_{k}\Big] \mathrm{d}z, \\
    &=\int q_{j} \prod_{k \neq j } q_{k} \Bigg[ \log q_{j} + \log \prod_{k \neq j } q_{k}\Bigg] \mathrm{d}z, \\
    &=\int \Big[ q_{j} \log q_{j} \prod_{k \neq j } q_{k} + q \log \prod_{k \neq j } q_{k}\Big] \mathrm{d}z, \\
    &=\int \int \Big[ q_{j} \log q_{j} \prod_{k \neq j } q_{k} + q \log \prod_{k \neq j } q_{k}\Big] \mathrm{d}z_{j} \mathrm{d}z_{k \neq j}, \\
    &=\int q_{j} \log q_{j} \Bigg[\int \prod_{k \neq j } q_{k} \mathrm{d}z_{k \neq j} \Bigg]\mathrm{d}z_{j} + \int \int q \log \prod_{k \neq j } q_{k} \mathrm{d}z_{j} \mathrm{d}z_{k \neq j} \\
    &=\int q_{j} \log q_{j} \mathrm{d}z_{j} + \int  \prod_{k \neq j } q_{k} \log \prod_{k \neq j } q_{k}\int q_{j} \mathrm{d}z_{j}  \mathrm{d}z_{k \neq j},\\
    &=\int q_{j} \log q_{j} \mathrm{d}z_{j} + \int  \prod_{k \neq j } q_{k} \log \prod_{k \neq j } q_{k} \mathrm{d}z_{k \neq j},\\
    &=\int q_{j} \log q_{j}(z_{j}) \mathrm{d}z_{j} + \mathcal{H}_{k \neq j},
\end{align*}
\end{subequations}
where
\begin{equation*}
    \mathcal{H}_{k \neq j} = \int  \prod_{k \neq j } q_{k} \log \prod_{k \neq j } q_{k} \mathrm{d}z_{k \neq j},
\end{equation*}
which is the entropy of all the factorized probability densities $k \neq j$.

\section{Fisher and Symmetrized KL-divergence}
\label{appendix:appendix_B}

Given a probability density function $q(z; \xi)$, the symmetrized KL-divergence captures the movement of a distance of $\Delta\xi$ in the direction of the steepest ascent as the dissimilarity of the probability densities $q(z; \xi)$ and $q(z; \xi + \Delta\xi)$, and is formulated as:
\begin{equation}
    \label{equation:kld_sym}
    D^{\text{sym}}_{\text{KL}}(\xi,\xi+\Delta\xi) = \mathds{E}_{q(z; \xi)}\bigg[\log\frac{q(z; \xi)}{q(z; \xi+\Delta\xi)}\bigg] + \mathds{E}_{q(z; \xi + \Delta\xi)}\biggl[\log\frac{q(z; \xi+\Delta\xi)}{q(z; \xi)} \bigg].
\end{equation}
Additionally, the second order Taylor expansion for a function $f(\xi)$ at a point $\xi_{i}$ is given by:
\begin{equation}
\label{equation:taylor_2nd}
    f(\xi) \approx f(\xi_{i}) + \nabla_{\xi}f(\xi_{i})^T(\xi - \xi_{i}) + \frac{1}{2}(\xi - \xi_{i})^T\nabla_{\xi}^2f(\xi_{i})(\xi - \xi_{i}).
\end{equation}
Substituting $\xi = \xi_{i} + \Delta\xi$ (such that $\Delta\xi \rightarrow 0$) in Equation \ref{equation:taylor_2nd}, we get:
\begin{equation}
\label{equation:taylor_final}
   f(\xi_{i} + \Delta\xi) = f(\xi_{i}) + \nabla_{\xi}f(\xi_{i})^T\Delta\xi + \frac{1}{2}\Delta\xi^T\nabla_{\xi}^2f(\xi_{i})\Delta\xi. 
\end{equation}
Now, using the result in Equation \ref{equation:taylor_final} we can expand the terms $\log q(z; \xi+\Delta\xi)$ from the right hand side of Equation \ref{equation:kld_sym} as follows:
\begin{equation}
\label{equation:expand}
    \log q(z; \xi+\Delta\xi) = \log q(z; \xi) + \{\nabla_{\xi}\log q(z; \xi)\}^T\Delta\xi + \frac{1}{2}\Delta\xi^T\nabla_{\xi}^2\log q(z; \xi)\Delta\xi.
\end{equation}
Using the result from Equation \ref{equation:expand} we expand the first term on the right-hand side of Equation \ref{equation:kld_sym} as:
\begin{eqnarray}
\label{equation:kl_fisher}
    \log\frac{q(z; \xi)}{q(z; \xi+\Delta\xi)} &=& \log q(z; \xi) - \log q(z; \xi+\Delta\xi) \nonumber \\
    &=&-\{\nabla_{\xi}\log q(z; \xi)\}^T\Delta\xi - \frac{1}{2}\Delta\xi^T\nabla_{\xi}^2\log q(z; \xi)\Delta\xi, \nonumber \\
    &=&-\frac{\{\nabla_{\xi} q(z; \xi)\}^T \Delta\xi}{q(z; \xi)} - \frac{1}{2}\Delta\xi^T\nabla_{\xi}\bigg\{\frac{\nabla_{\xi} q(z; \xi)}{q(z; \xi)}\bigg\}\Delta\xi \nonumber, \\
    &=&-\frac{\{\nabla_{\xi} q(z; \xi)\}^T \Delta\xi}{q(z; \xi)} - \frac{1}{2}\Delta\xi^T\bigg\{\frac{q(z; \xi)\nabla_{\xi}^2 q(z; \xi)}{q(z; \xi)q(z; \xi)} \nonumber\\
    & &-\frac{\nabla_{\xi}q(z; \xi)\nabla_{\xi}q(z; \xi)^T}{q(z; \xi)q(z; \xi)} \bigg\} \Delta\xi, \nonumber \\
    &=&-\frac{\{\nabla_{\xi} q(z; \xi)\}^T \Delta\xi}{q(z; \xi)} - \frac{1}{2}\Delta\xi^T\bigg\{ \frac{\nabla_{\xi}^2 q(z; \xi)}{q(z; \xi)}\bigg\} \Delta\xi \nonumber \\ 
    & &+\frac{1}{2}\Delta\xi^T\bigg\{\nabla_{\xi} \log q(z|x; \xi) \nabla_{\xi} \log q(z|x; \xi)^T\bigg\} \Delta\xi.
\end{eqnarray}
Taking expectations with respect to $q(z; \xi)$ on both sides of Equation \ref{equation:kl_fisher}, we get:
\begin{eqnarray}
\label{equation:kld_fisher_int}
    \mathds{E}_{q(z; \xi)}\bigg[\log\frac{q(z; \xi)}{q(z; \xi+\Delta\xi)}\bigg] 
    &=& -\mathds{E}_{q(z; \xi)}\bigg[\frac{\{\nabla_{\xi} q(z; \xi)\}^T \Delta\xi}{q(z; \xi)}\bigg] - \mathds{E}_{q(z; \xi)}\bigg[\frac{1}{2}\Delta\xi^T\bigg\{ \frac{\nabla_{\xi}^2 q(z; \xi)}{q(z; \xi)}\bigg\} \Delta\xi\bigg] \nonumber\\
    & &+ \mathds{E}_{q(z; \xi)}\bigg[\frac{1}{2}\Delta\xi^T\bigg\{\nabla_{\xi} \log q(z|x; \xi) \nabla_{\xi} \log q(z|x; \xi)^T\bigg\} \Delta\xi\bigg], \nonumber \\
    &=&-\bigg[\int q(z; \xi)\frac{\{\nabla_{\xi} q(z; \xi)\}^T}{q(z; \xi)}\mathrm{d}z\bigg]\Delta\xi \nonumber\\
    & &- \frac{1}{2}\Delta\xi^T\bigg[ \int q(z; \xi)\bigg\{ \frac{\nabla_{\xi}^2 q(z; \xi)}{q(z; \xi)}\mathrm{d}z\bigg\} \bigg]\Delta\xi \nonumber\\ 
    & &+ \frac{1}{2}\Delta\xi^T\bigg\{\mathds{E}_{q(z; \xi)}\bigg[ \nabla_{\xi} \log q(z|x; \xi) \nabla_{\xi} \log q(z|x; \xi)^T \bigg] \bigg\}\Delta\xi.
\end{eqnarray}
The terms on the right-hand side of Equation \ref{equation:kld_fisher_int} can each be evaluated as:
\begin{align}
\label{equation:1st}
    \bigg[\int q(z; \xi)\frac{\{\nabla_{\xi} q(z; \xi)\}^T}{q(z; \xi)}\mathrm{d}z\bigg]\Delta\xi &=\bigg[\int \{\nabla_{\xi} q(z; \xi)\}^T\mathrm{d}z\bigg]\Delta\xi \nonumber \\
    &=\bigg[\{\nabla_{\xi} \int q(z; \xi)\mathrm{d}z\}^T\bigg]\Delta\xi \nonumber \\
    &=\bigg[\{\nabla_{\xi} 1\}^T \bigg]\Delta\xi \nonumber \\
    &=0,
\end{align}
\begin{align}
\label{equation:2nd}
    \frac{1}{2}\Delta\xi^T\bigg[ \int q(z; \xi)\bigg\{ \frac{\nabla_{\xi}^2 q(z; \xi)}{q(z; \xi)}\mathrm{d}z\bigg\} \bigg]\Delta\xi &=  \frac{1}{2}\Delta\xi^T\bigg[ \int \nabla_{\xi}^2 q(z; \xi)\mathrm{d}z \bigg]\Delta\xi \nonumber \\
    &=\frac{1}{2}\Delta\xi^T\bigg[ \nabla_{\xi}^2 \bigg\{\int q(z; \xi)\mathrm{d}z\bigg\} \bigg]\Delta\xi \nonumber \\
    &=\frac{1}{2}\Delta\xi^T\bigg[ \nabla_{\xi}^2 1 \bigg]\Delta\xi \nonumber \\
    &= 0,
\end{align}
and
\begin{equation}
\label{equation:3rd}
   \frac{1}{2}\Delta\xi^T\bigg\{\mathds{E}_{q(z; \xi)}\bigg[ \nabla_{\xi} \log q(z|x; \xi) \nabla_{\xi} \log q(z|x; \xi)^T \bigg] \bigg\}\Delta\xi  = \frac{1}{2}\Delta\xi^T I(\xi) \Delta\xi,
\end{equation}
where $I(\xi)$ is the Fisher Information matrix.

Thus substituting the results from Equations \ref{equation:1st}, \ref{equation:2nd}, and \ref{equation:3rd} in Equation \ref{equation:kld_fisher_int}, we get:
\begin{equation}
\label{equation:kl_fisher_1}
    \mathds{E}_{q(z; \xi)}\bigg[\log\frac{q(z; \xi)}{q(z; \xi+\Delta\xi)}\bigg] = \frac{1}{2}\Delta\xi^T I(\xi)\Delta\xi.
\end{equation}
As $\Delta\xi \rightarrow 0$, therefore,  $q(z; \xi)$ and $q(z; \xi+\Delta\xi)$ are the same probability density. 
Thus, expanding the second term on the right-hand side of Equation \ref{equation:kld_sym}, in a similar manner, we can show that:
\begin{equation}
\label{equation:kl_fisher_2}
    \mathds{E}_{q(z; \xi + \Delta\xi)}\biggl[\log\frac{q(z; \xi+\Delta\xi)}{q(z; \xi)} \bigg] = \frac{1}{2}\Delta\xi^T I(\xi)\Delta\xi.
\end{equation}
Combining the results from Equations \ref{equation:kl_fisher_1} and \ref{equation:kl_fisher_2} in Equation \ref{equation:kld_sym} as follows:
\begin{align}
    D^{\text{sym}}_{\text{KL}}(\xi,\xi+\Delta\xi) &\approx \frac{1}{2}\Delta\xi^T I(\xi)\Delta\xi + \frac{1}{2}\Delta\xi^T I(\xi)\Delta\xi \nonumber \\
    &\approx \Delta\xi^T I(\xi)\Delta\xi,
\end{align}
which corresponds to computing the inner product of a vector with itself in the Riemannian manifold \shortcite{chen2018metrics}.

\section{Fisher and Hessian}
\label{appendix:appendix_C}

Given a function $f(x)$, its Jacobian $\mathbb{J}[f(x)]$ and Hessian $H[f(x)]$ are computed as
\begin{equation*}
    H[f(x)] = \mathbb{J}[\nabla f(x)].
\end{equation*}
For a probability density $q(z|x_{i}; \xi_{i})$, the Hessian for $\log q(z|x_{i}; \xi_{i})$ is given by:
\begin{align*}
    H[\log q(z|x_{i}; \xi_{i})] &= \mathbb{J}\bigg[ \nabla_{\xi_{i}} \log q(z|x_{i}; \xi_{i}) \bigg], \\
    &= \mathbb{J}\bigg[ \frac{\nabla_{\xi_{i}} q(z|x_{i}; \xi_{i})}{q(z|x_{i}; \xi_{i})}\bigg], \\
    &= \nabla_{\xi_{i}}\bigg[ \frac{\nabla_{\xi_{i}} q(z|x_{i}; \xi_{i})}{q(z|x_{i}; \xi_{i})} \bigg], \\
    &= \frac{q(z|x_{i}; \xi_{i})H[q(z|x_{i}; \xi_{i})] - \nabla_{\xi_{i}} q(z|x_{i}; \xi_{i})\nabla_{\xi_{i}} q(z|x_{i}; \xi_{i})^T}{q(z|x_{i}; \xi_{i})q(z|x_{i}; \xi_{i})}, \\
    &= \frac{H[q(z|x_{i}; \xi_{i})]}{q(z|x_{i}; \xi_{i})} - \bigg(\frac{\nabla_{\xi_{i}} q(z|x_{i}; \xi_{i})}{q(z|x_{i}; \xi_{i}} \bigg) \bigg(\frac{\nabla_{\xi_{i}} q(z|x_{i}; \xi_{i})}{q(z|x_{i}; \xi_{i})} \bigg)^T, \\
    &= \frac{H[q(z|x_{i}; \xi_{i})]}{q(z|x_{i}; \xi_{i})} - \nabla_{\xi_{i}} \log q(z|x_{i}; \xi_{i}) \nabla_{\xi_{i}} \log q(z|x_{i}; \xi_{i})^T.
\end{align*}
Therefore, taking expectations with respect to $q(z|x_{i}; \xi_{i})$, we get:
\begin{align*}
    \mathds{E}_{q}\bigg[ H[\log q(z|x_{i}; \xi_{i})] \bigg]  &= \mathds{E}_{q}\bigg[\frac{H[q(z|x_{i}; \xi_{i})]}{q(z|x_{i}; \xi_{i})}\bigg] - \mathds{E}_{q}\bigg[\nabla_{\xi_{i}} \log q(z|x_{i}; \xi_{i}) \nabla_{\xi_{i}} \log q(z|x_{i}; \xi_{i})^T\bigg], \\
    &= \int \frac{H[q(z|x_{i}; \xi_{i})]}{q(z|x_{i}; \xi_{i})} q(z|x_{i}; \xi_{i}) \mathrm{d}z- I(\xi_{i}), \\
    &= \int H[q(z|x_{i}; \xi_{i})]\mathrm{d}z- I(\xi_{i}), \\
    &= \mathbb{J} \bigg[ \int \nabla_{\xi_{i}} q(z|x_{i}; \xi_{i}) \mathrm{d}z\bigg] - I(\xi_{i}), \\
    &= \mathbb{J} \bigg[ \nabla_{\xi_{i}} \int q(z|x_{i}; \xi_{i}) \mathrm{d}z\bigg] - I(\xi_{i}), \\
    &= \mathbb{J} \bigg[ \nabla_{\xi_{i}} 1 \bigg] - I(\xi_{i}), \\
    &= - I(\xi_{i}), \\
    I(\xi_{i}) &= - \mathds{E}_{q}\bigg[ H[\log q(z|x_{i}; \xi_{i})] \bigg].
\end{align*}

\vskip 0.2in
\bibliography{jair_references}
\bibliographystyle{theapa}

\end{document}